\documentclass[runningheads]{llncs}

\usepackage{eccv}

\usepackage{eccvabbrv}

\usepackage{graphicx}
\usepackage{booktabs}
\usepackage{multirow}
\usepackage{url}
\usepackage[accsupp]{axessibility}  % Improves PDF readability for those with disabilities.

\usepackage{hyperref}

\usepackage{orcidlink}

\begin{document}

\title{Enriching Information and Preserving Semantic Consistency in Expanding Curvilinear Object Segmentation Datasets} 

\titlerunning{Expanding Curvilinear Object Segmentation Datasets}

\author{Qin Lei\orcidlink{0000-0002-1340-5969} \and
Jiang Zhong\orcidlink{0000-0002-5169-4634} \and
Qizhu Dai\orcidlink{0000-0003-1072-7847}}

\authorrunning{Qin Lei et al.}

\institute{School of Computer Science, Chongqing University, Chongqing 400044, China \\
\email{\{qinlei,zhongjiang,daiqizhu\}@cqu.edu.cn}}

\maketitle

\begin{abstract}
  Curvilinear object segmentation plays a crucial role across various applications, yet datasets in this domain often suffer from small scale due to the high costs associated with data acquisition and annotation. To address these challenges, this paper introduces a novel approach for expanding curvilinear object segmentation datasets, focusing on enhancing the informativeness of generated data and the consistency between semantic maps and generated images.
  Our method enriches synthetic data informativeness by generating curvilinear objects through their multiple textual features. By combining textual features from each sample in original dataset, we obtain synthetic images that beyond the original dataset's distribution. This initiative necessitated the creation of the Curvilinear Object Segmentation based on Text Generation (COSTG) dataset. Designed to surpass the limitations of conventional datasets, COSTG incorporates not only standard semantic maps but also some textual descriptions of curvilinear object features.
  To ensure consistency between synthetic semantic maps and images, we introduce the Semantic Consistency Preserving ControlNet (SCP ControlNet). This involves an adaptation of ControlNet with Spatially-Adaptive Normalization (SPADE), allowing it to preserve semantic information that would typically be washed away in normalization layers. This modification facilitates more accurate semantic image synthesis.
  Experimental results demonstrate the efficacy of our approach across three types of curvilinear objects (angiography, crack and retina) and six public datasets (CHUAC, XCAD, DCA1, DRIVE, CHASEDB1 and Crack500). The synthetic data generated by our method not only expand the dataset, but also effectively improves the performance of other curvilinear object segmentation models. Source code and dataset are available at \url{https://github.com/tanlei0/COSTG}.
\end{abstract}

\section{Introduction}
\label{sec:intro}
Segmentation of curvilinear objects is crucial for a wide range of applications. These applications include assessing road conditions and their maintenance\cite{lei2024joint, lei2024integrating, lei2023dynamic}, screening for retinal fundus diseases\cite{abramoff2010retinal, fraz2012blood, lei2023adaptive} and diagnosing coronary artery diseases\cite{thomas2017novel}. Despite extensive research in the literature, accurately segmenting curvilinear objects remains a challenge. Their thin, long, and tortuous shapes, along with numerous tiny branches, contribute to their complex structure. Additionally, the boundaries of curvilinear objects are often blurred due to imaging conditions and noisy backgrounds.

Current deep supervised learning approaches for curvilinear object segmentation, while promising, heavily rely on extensive pixel-wise manual annotations for training\cite{wang2019context, cheng2021joint, shi2022local,shit2021cldice}. 
This requirement presents a significant challenge in the domain of curvilinear object segmentation for several reasons. Firstly, the intricate nature of curvilinear objects, characterized by their thin, elongated forms and complex branching, makes accurate annotation both difficult and time-consuming. Secondly, the quality of the segmentation is often compromised by factors such as poor image resolution, inconsistency between annotations and images. Additionally, privacy concerns in certain applications, like road condition assessment and coronary artery disease diagnosis, restrict access to large-scale, annotated datasets. Although a few publicly available datasets\cite{DRIVEstaal2004ridge, XCADma2021self, CHASE_CHUAC2018, DCA1cervantes2019automatic, Crack500yang2019feature} exist, they are typically small, ranging from dozens to a few hundred samples, which is insufficient by current deep learning standards. This scarcity of large, accurately annotated datasets hampers the scaling and improvement of supervised segmentation models in this field.

To mitigate the aforementioned challenges, expanding the dataset has emerged as a viable solution. Beyond traditional data augmentation techniques\cite{cubuk2020randaugment, hendrycks2019augmix}, the advancement of deep generative models has shown considerable success in dataset expansion, especially in image classification tasks\cite{yuan2023real,zhang2023expanding,azizi2023synthetic}. A key strategy for expanding image segmentation datasets with deep generative models involves first obtaining semantic maps through various means. Subsequently, these semantic maps are used to generate images that are precisely paired with them, employing conditional generation techniques\cite{jin2023establishment,chai2022synthetic}.

However, the approach of expanding image segmentation datasets using semantic maps for conditional image generation confronts two principal challenges. 
The first pertains to whether the synthetic images introduce sufficient new information to beyond the original dataset's distribution\cite{zhang2023expanding}. 
Larger sacle training that do not surpass the original distribution risk leading to overfitting, thereby reducing the model's generalizability and performance. 
While traditional data augmentation techniques can somewhat beyond the original distribution\cite{cubuk2020randaugment, hendrycks2019augmix}, deep generative models are particularly prone to this pitfall\cite{zhang2023expanding}. 
This has been underscored by previous studies, which demonstrated that models trained exclusively on synthetic data underperform compared to those trained on original data, however a combination of both can improve performance\cite{yuan2023real,zhang2023expanding,azizi2023synthetic}.

The second challenge centers on maintaining a high degree of consistency between semantic maps and synthetic images. While the Generative Adversarial Networks (GAN)\cite{goodfellow2014generative} era marked substantial progress in achieving precise semantic image synthesis\cite{park2019semantic,tan2021efficient}, the transition to the Diffusion Models era\cite{rombach2022high} introduces additional complexities. 
The integration of more conditions\cite{rombach2022high,zhang2023adding,mou2023t2i}, notably text, into the image generation process brings forth the challenge of ensuring that semantic information is not "washed away" by normalization layers\cite{park2019semantic}. 
Overcoming this obstacle is crucial for attaining accurate semantic image synthesis\cite{park2019semantic,tan2021efficient}.

\begin{figure}[!ht]
	\centering
	\includegraphics[width=1\textwidth]{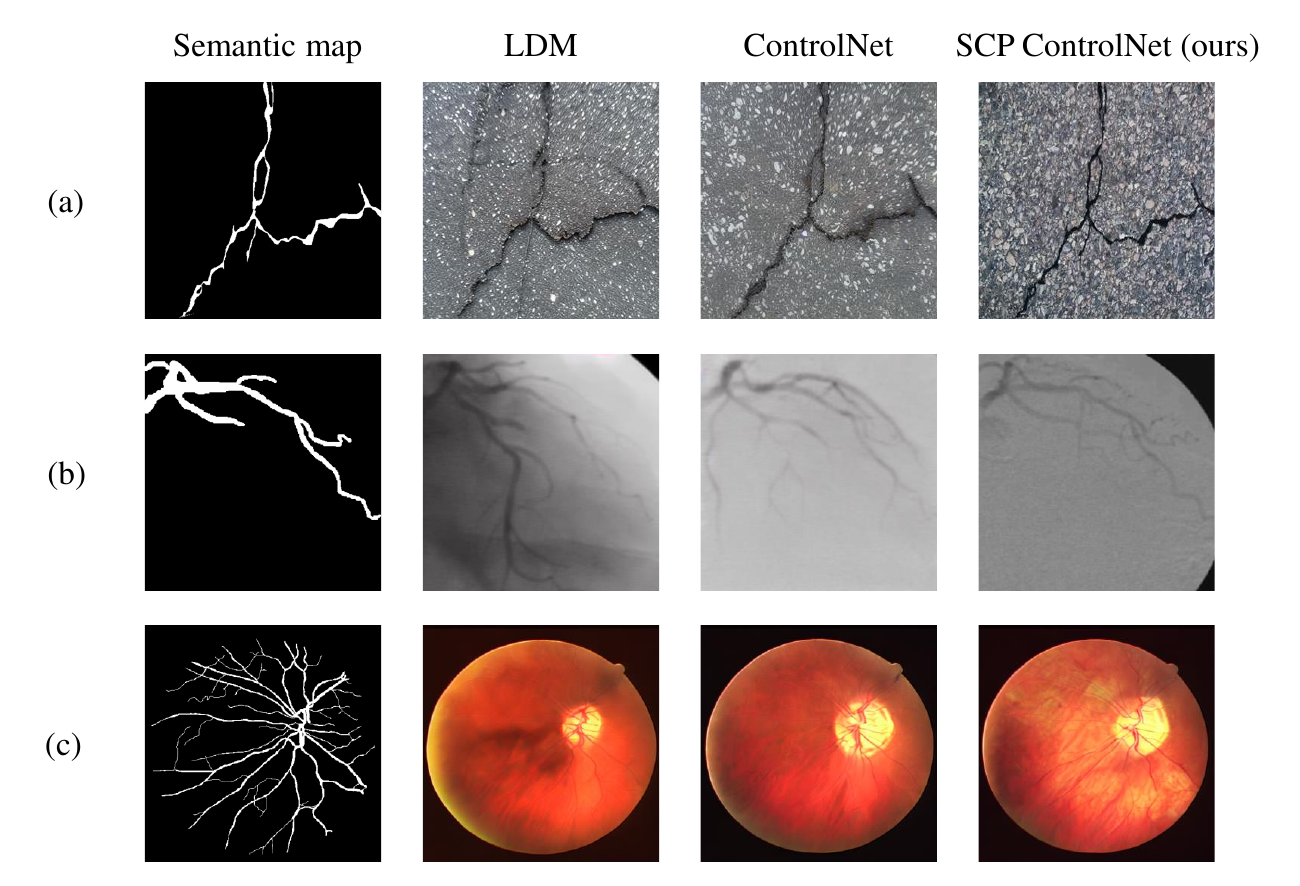}
	\caption{Comparative analysis of curvilinear objects semantic image synthesis among SCP ControlNet, LDM, and ControlNet, where (a), (b), and (c) illustrate the three types of curvilinear objects in this paper: (a) crack, (b) angiography and (c) retina. It demonstrates SCP ControlNet's enhanced precision in reflecting semantic map information within generated images, crucial for expanding semantic segmentation datasets.}
	\label{fig:compare_1}
\end{figure}

To extend the knowledge beyond the original dataset's distribution, this paper introduces a method that combines multiple textual features of curvilinear objects to generate semantic maps. 
These semantic maps significantly beyond the distribution of the original dataset. 
To implement this method, we developed the Curvilinear Object Segmentation based on Text Generation (COSTG) dataset.
COSTG is designed to surpass the limitations of traditional datasets by including not only standard semantic maps but also textual descriptions of curvilinear object features.
To achieve more accurate semantic image synthesis and ensure consistency between synthetic semantic maps and images, 
we introduce the Semantic Consistency Preserving ControlNet (SCP ControlNet).
This involves an adaptation of ControlNet\cite{zhang2023adding} with Spatially-Adaptive Normalization (SPADE)\cite{park2019semantic}, allowing it to preserve semantic information that would typically be washed away in normalization layers\cite{park2019semantic}.

\cref{fig:compare_1} underscores the efficacy of SCP ControlNet over LDM and ControlNet in generating images that more accurately embody the details of semantic maps, particularly for curvilinear object tasks such as (a) crack, (b) angiography, and (c) retina. 
The ability of SCP ControlNet to preserving a high degree of semantic consistency between semantic maps and generated images is crucial for expanding semantic segmentation datasets. 
For more details about \cref{fig:compare_1} such as prompts please refer to the Appendix.

Leveraging the COSTG dataset, we fine-tuned Stable Diffusion 1.5\cite{rombach2022high} and our SCP ControlNet. The effectiveness of our method is evidenced by experiments conducted on expanded datasets across three types of curvilinear objects (angiography, cracks, and retina) and six public datasets (CHUAC\cite{CHASE_CHUAC2018}, XCAD\cite{XCADma2021self}, DCA1\cite{DCA1cervantes2019automatic}, DRIVE\cite{DRIVEstaal2004ridge}, CHASEDB1\cite{CHASE_CHUAC2018}, and Crack500\cite{Crack500yang2019feature}).
These experiments demonstrate that our approach can achieve significant performance improvements with fewer generated samples, underscoring its efficiency and potential in enhancing the segmentation of curvilinear objects.

The contributions of this paper are summarized as follows:
\begin{itemize}
	\item We propose a method capable of generating images that beyond the distribution of the original dataset. This is achieved by combining multiple textual features of curvilinear objects as conditions for generation.
	\item To facilitate the generation of curvilinear objects with conditions based on textual features, we created the Curvilinear Object Segmentation based on Text Generation (COSTG) dataset. This dataset incorporates not only conventional semantic maps but also textual descriptions detailing multiple features of curvilinear objects.
	\item We present the Semantic Consistency Preserving ControlNet (SCP ControlNet), an adaptation of ControlNet with Spatially-Adaptive Normalization. This enhancement allows ControlNet to retain semantic information usually lost during normalization, leading to more precise semantic image synthesis.
\end{itemize}

\section{Related Work}
\subsection{Expanding dataset}
In the field of computer vision, beyond collecting more data, there are generally two approaches to expanding datasets: traditional data augmentation and generative data augmentation based on generative models.

Traditional image data augmentation has become a primary means to enhance the generalization ability of DNNs during model training processes\cite{shorten2019survey,yang2022image}.
Depending on the technique, image data augmentation can be categorized into four main types: image manipulation, image erasing, image mix, and auto augmentation. 
Image manipulation enhances data through random transformations such as flipping, rotating, scaling, cropping, sharpening, and translation\cite{yang2022image}.
On the other hand, image erasing \cite{devries2017improved,zhong2020random,chen2020gridmask,li2020fencemask} replaces pixels in certain image areas with constant or random values. 
Image mix combines two or more images or subregions into a single image\cite{zhang2020does,yun2019cutmix,hendrycks2019augmix}. Lastly, auto augmentation determines augmentation operations from a set of random enhancements using search algorithms or random selection\cite{cubuk2019autoaugment,lim2019fast,cubuk2020randaugment}. 
Although these methods have shown effectiveness in certain applications, they primarily enhance data by applying predefined transformations to each image.
This leads only to local variations in pixel values and does not generate images with significantly diversified content. 
Moreover, as most methods employ random operations, they cannot ensure that the enhanced samples are informative for model training and may even introduce noisy samples.
Therefore, the new information brought is often insufficient to expand small-scale datasets, resulting in inefficient expansion.

In the past two years, generative data augmentation has been indispensably dominated by GANs\cite{li2022bigdatasetgan, he2022synthetic}. 
With the development of diffusion models, compared to GAN models, diffusion models offer higher fidelity, more stable training, and easier controllability advantages\cite{ho2020denoising, zhang2023adding}. Especially in the field of image classification, an increasing number of studies based on diffusion models for generative data augmentation have emerged\cite{he2022synthetic,trabucco2023effective,saharia2022photorealistic,azizi2023synthetic}, as using categories as conditions for generation is particularly natural and straightforward. 
The results of these studies mostly show that models trained solely on generated data do not perform as well as those trained solely on original data, but a combination of both can enhance performance\cite{yuan2023real,zhang2023expanding,azizi2023synthetic}. This is partly because the generated data does not surpass the original data distribution, leading to increased overfitting and decreased generalization performance\cite{zhang2023expanding}. On the other hand, the generated images carry a certain level of noise, leading to a decrease in the model's fitting ability\cite{yuan2023real}.

Besides image classification tasks, there are currently two methods of generative data augmentation for image segmentation tasks.
One involves generating images conditioned on semantic maps~\cite{giambi2023conditioning,fontanini2023semantic}. 
The other achieves the generation of both semantic maps and images through multimodal attention maps~\cite{wu2023diffumask}. 
In expanding datasets for curvilinear object segmentation, the approach of generating images conditioned on semantic maps is commonly used\cite{chai2022synthetic,jin2023establishment}. 
Although this paper proposes curvilinear object generation based on text, the attention maps between text and the curvilinear objects in images do not correspond one-to-one, making it difficult to simultaneously generate semantic maps and images through attention maps.
\subsection{Diffusion models}
Diffusion models represent a category of generative models designed to reverse a stochastic process that incrementally degrades data with Gaussian noise\cite{ho2020denoising,sohl2015deep}. 
This process encompasses two distinct phases: a forward phase where noise is added to the data progressively from time \(t = 0\) to \(t = T\), and a reverse phase which aims to remove this noise, transitioning from \(t = T\) back to \(t = 0\). The denoising, or reverse phase, involves sampling from a Gaussian distribution \(\mathcal{N}(x_{t-1}; \mu_\theta(x_t, t), \Sigma_\theta(x_t, t))\), where the mean \(\mu_\theta(x_t, t)\) is conditioned on the noisy sample \(x_t\) at the subsequent time step, and the variance \(\Sigma_\theta(x_t, t)\) is dictated by a pre-established schedule.

Distinguished from traditional diffusion models, Latent Diffusion Models (LDMs)\cite{rombach2022high} map the original high-dimensional data \(x_0\) to a lower-dimensional latent space \(z_0\), conducting the diffusion process within this latent realm. This technique notably enhances sampling efficiency. LDMs utilize an autoencoder \((E, D)\) configuration, encoding \(x_0\) into \(z_0 = E(x_0)\) and subsequently decoding \(z_0\) back to \(\tilde{x_0} = D(z_0)\). Within the latent space, the diffusion equation is given by \(z_t = \sqrt{\beta_t}z_{t-1} + \sqrt{1-\beta_t}\epsilon_t\), and the reverse process is modeled as \(p_\theta(z_{t-1}|z_t) = \mathcal{N}(z_{t-1}; \mu_{\theta}(z_t, t), \Sigma_\theta(t))\), where \(\mu_{\theta}(z_t, t)\) is a neural network designed to reconstruct the preceding latent vector from its successor, and \(\beta\) denotes a noise variance schedule that regulates the level of noise added or subtracted at each timestep. Here, \(p_\theta\) represents the conditional probability distribution of the data or latent vector at the previous timestep given the noisy sample at the next timestep.

\subsection{Semantic image synthesis within diffusion models}
Semantic image synthesis within diffusion models can be delineated into two distinct phases. Prior to methods represented by Stable Diffusion (SD) and Latent Diffusion Models (LDMs)\cite{rombach2022high}, semantic image synthesis typically revolved around using semantic maps as the sole condition, as exemplified by seminal works like SDM\cite{wang2022semantic}. 
Some studies ventured further by incorporating additional encoded conditions into the semantic map channels, such as attributes of categories within images\cite{tan2021efficient} or facial IDs\cite{giambi2023conditioning}.

After SD, methods based on LDMs have become paradigmatic, integrating textual features as part of the cross-attention mechanism (as the key (K) and value (V) components of QKV), considering the influence of text information during the conditional generation process.
To infuse more conditions into SD, methodologies such as ControlNet\cite{zhang2023adding}, T2I-adapter\cite{mou2023t2i}, Composer\cite{huang2023composer} and Uni-ControlNet\cite{zhao2024uni} have introduced extra network structures to realize more complex conditional generations, building upon the foundation where textual features form part of the cross-attention.

However, ControlNet and other LDM-based methods, when dealing with numerous conditions, can be overly simplistic, not taking into account the unique representational capabilities of each feature, such as bounding boxes\cite{fan2023frido}, complex scene graphs\cite{fan2023frido}, facial IDs\cite{wang2024instantid}, etc.
For instance, these general models, including LDM, ControlNet, and T2I-adapter, traditionally scale encoded semantic maps to the size of the latent noise and concatenate them directly to the latent noise for subsequent computations (Note: In LDM, semantic synthesis does not utilize textual features, while in ControlNet and T2I-adapter, textual features are incorporated as part of the cross-attention). 
Despite the introduction of additional network structures like ControlNet to utilize semantic information, this semantic information is still prone to being 'washed away' in normalization layers, leading to generated images that do not perfectly reflect the information in the semantic maps\cite{park2019semantic}.

Spatially adaptive normalization\cite{park2019semantic}, utilized in early SDM work, achieved highly accurate semantic image synthesis. However, SDM did not adopt the LDM paradigm by incorporating text features as part of the cross-attention mechanism and relied solely on semantic maps for precise semantic image synthesis. 
A significant contribution of this paper is the introduction of SCP ControlNet, which innovatively adapts ControlNet by the spatially adaptive normalization. 
This adaptation enables precise semantic image synthesis while concurrently utilizing textual information, aligning with the LDM framework.

\section{COSTG dataset}

Traditional curvilinear object segmentation datasets typically comprise only semantic maps and images of the curvilinear objects, which may not suffice for contemporary text-driven generative models, such as the increasingly popular Stable Diffusion (SD)\cite{rombach2022high}.
When it comes to expanding datasets, utilizing text as a condition can effectively introduce new information into the generated samples. 
This is particularly pertinent for small-scale curvilinear object segmentation datasets, where combining textual features can readily generate samples that surpass the original dataset's distribution.
To address this, we have developed the Curvilinear Object Segmentation based on Text Generation (COSTG) dataset.

\subsection{Dataset composition}
The COSTG dataset is constructed upon six existing public curvilinear object segmentation datasets: CHUAC\cite{CHASE_CHUAC2018}, XCAD\cite{XCADma2021self}, DCA1\cite{DCA1cervantes2019automatic}, DRIVE\cite{DRIVEstaal2004ridge}, CHASEDB1\cite{CHASE_CHUAC2018}, and Crack500\cite{Crack500yang2019feature}. 
Please refer to the Appendix for some examples of COSTG.

To enrich these datasets, we have appended textual descriptions encompassing six feature fields of the curvilinear objects.
The meanings of these six fields are as follows:
\begin{itemize}
	\item \textbf{Overview}: Provides a holistic description of the image. For semantic maps, it necessarily includes terms like "ground truth (GT)"  or "semantic map".
	\item \textbf{Dataset}: Specifies the public dataset from which the image is derived.
	\item \textbf{Location}: Describes the position of the curvilinear objects within the image.
	\item \textbf{Size}: Conveys the size and the proportion of the curvilinear objects within the image.
	\item \textbf{Trend and Shape}: Details the trend and shape of the curvilinear objects in the image.
	\item \textbf{Background}: Describes the background of the image, excluding the curvilinear objects. This field is omitted for semantic maps.
\end{itemize}

This structured and detailed composition of the COSTG dataset ensures that each image is accompanied by comprehensive textual information, enhancing the dataset's utility for text-driven generative modeling.

\subsection{Annotation process}
\begin{figure}[!ht]
	\centering
	\includegraphics[width=0.9\textwidth]{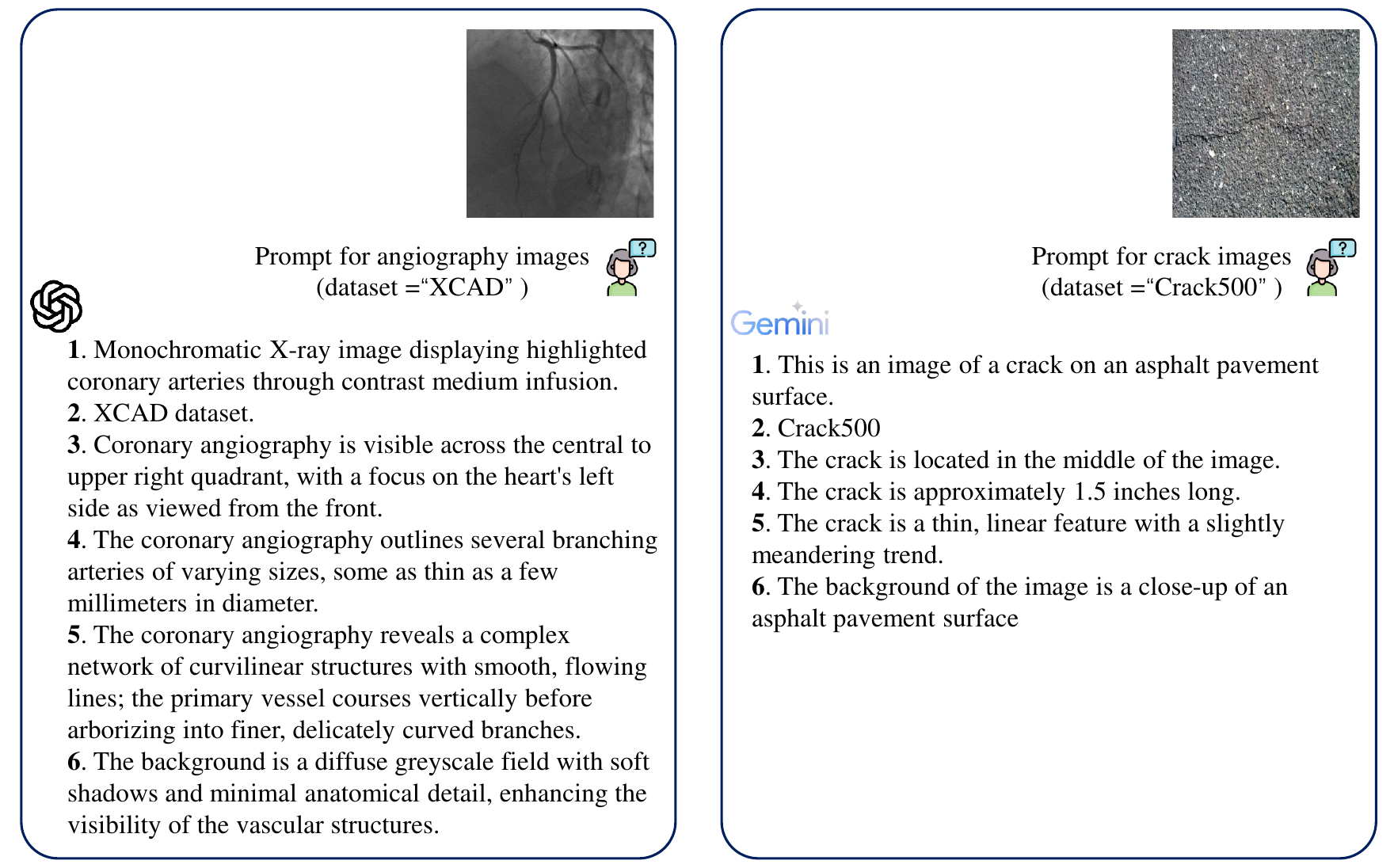}
	\caption{Two Examples of the Annotation Process: ChatGPT-4-V (Left) vs. Gemini 1.5 (Right).}
	\label{fig:LVM}
\end{figure}

All textual annotations within COSTG were performed by Language-Vision Models (LVMs). We evaluated multiple LVMs, including BLIP2\cite{li2023blip2}, Instruct BLIP\cite{instructblip}, LWM\cite{liu2023world}, ChatGPT-4-V\cite{achiam2023gpt}, and Gemini 1.5\cite{gemini}. Ultimately, we selected ChatGPT-4-V and Gemini 1.5, two large-scale LVMs, to complete the annotation process. \cref{fig:LVM} presents annotation examples using ChatGPT-4-V and Gemini 1.5. For more details on the annotation process, please refer to the Appendix.

\section{Methodology}
\subsection{SCP ControlNet}
\begin{figure}[!ht]
	\centering
	\includegraphics[width=0.95\textwidth]{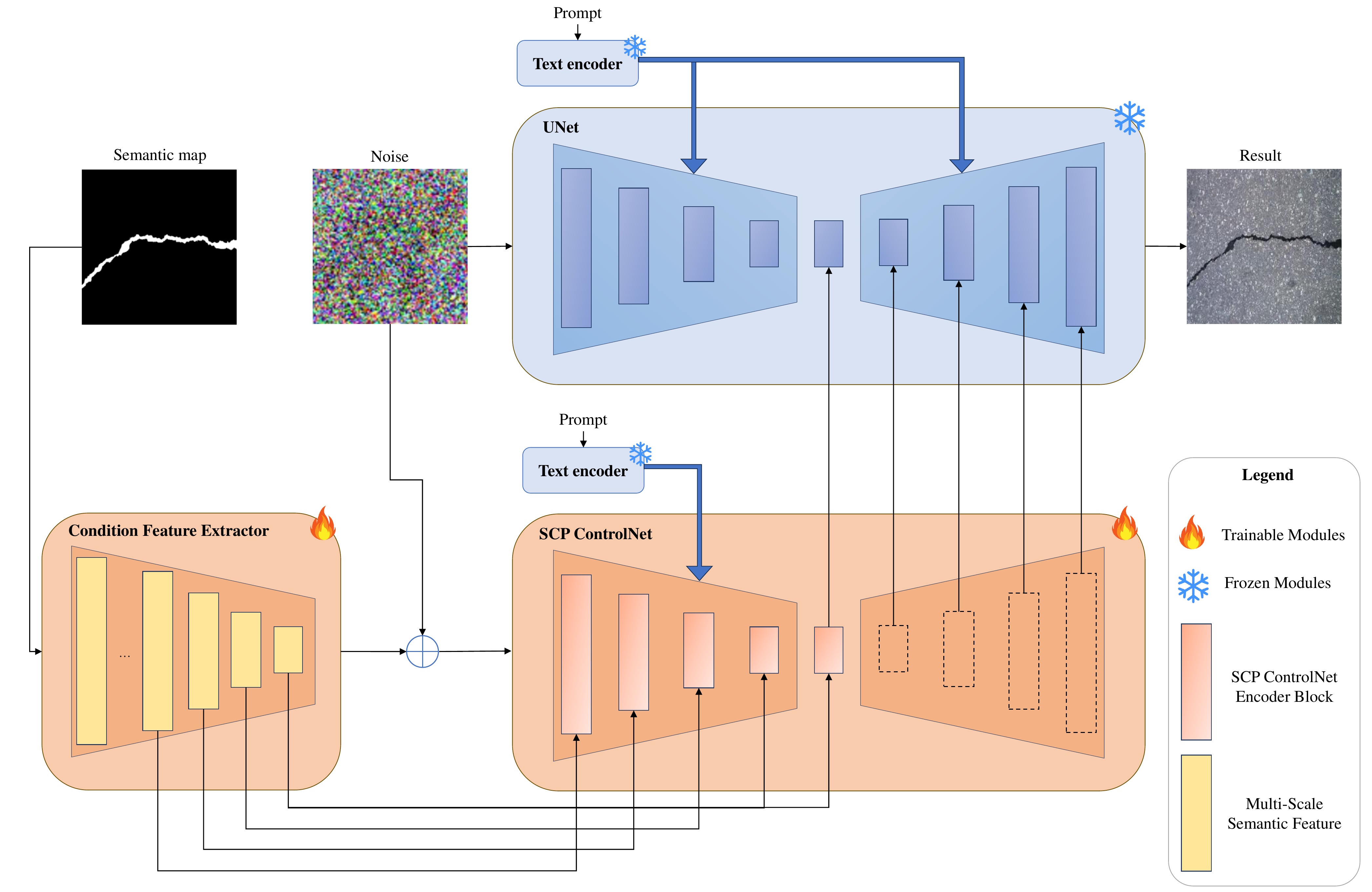}
	\caption{Overview of SCP ControlNet. Unlike ControlNet, SCP ControlNet features a distinct Encoder Block and employs a novel approach to utilizing semantic information. In addition to concatenating semantic information with latent noise, it infuses semantic data across multiple scales into various Encoder Blocks.}
	\label{fig:SCP_ControlNet}
\end{figure}

\begin{figure}[!ht]
	\centering
	\resizebox{0.5\textwidth}{!}{%
	\begin{subfigure}[b]{0.45\textwidth}
		\includegraphics[width=\textwidth]{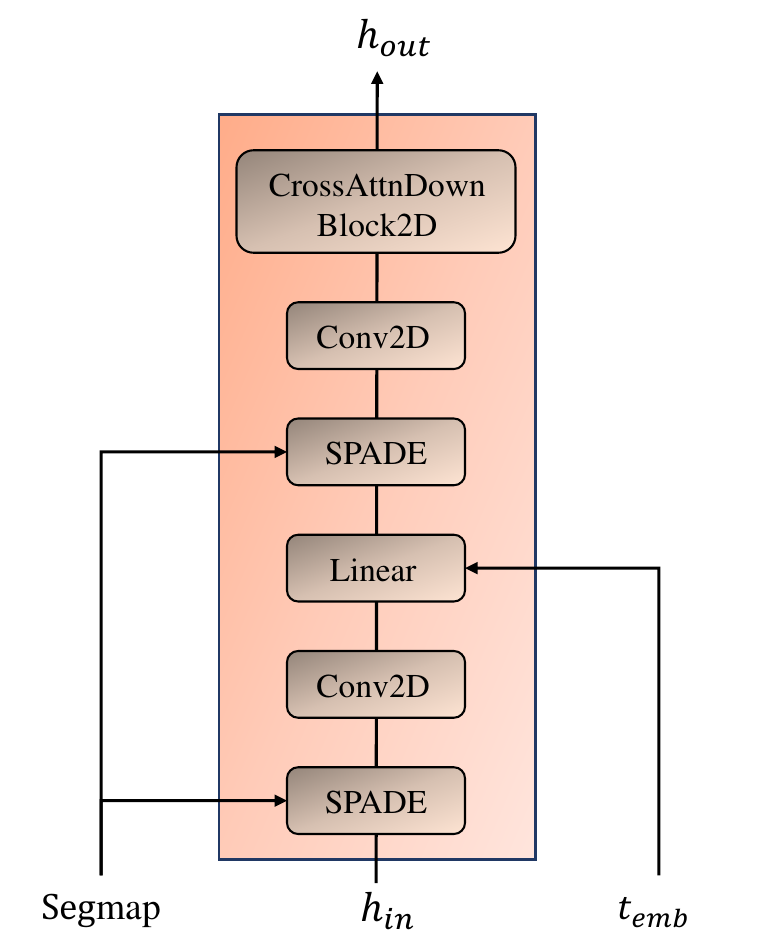}
		\caption{}
		\label{subfig:encoder}
	\end{subfigure}
	\hspace{20mm}
	\begin{subfigure}[b]{0.35\textwidth}
		\includegraphics[width=\textwidth]{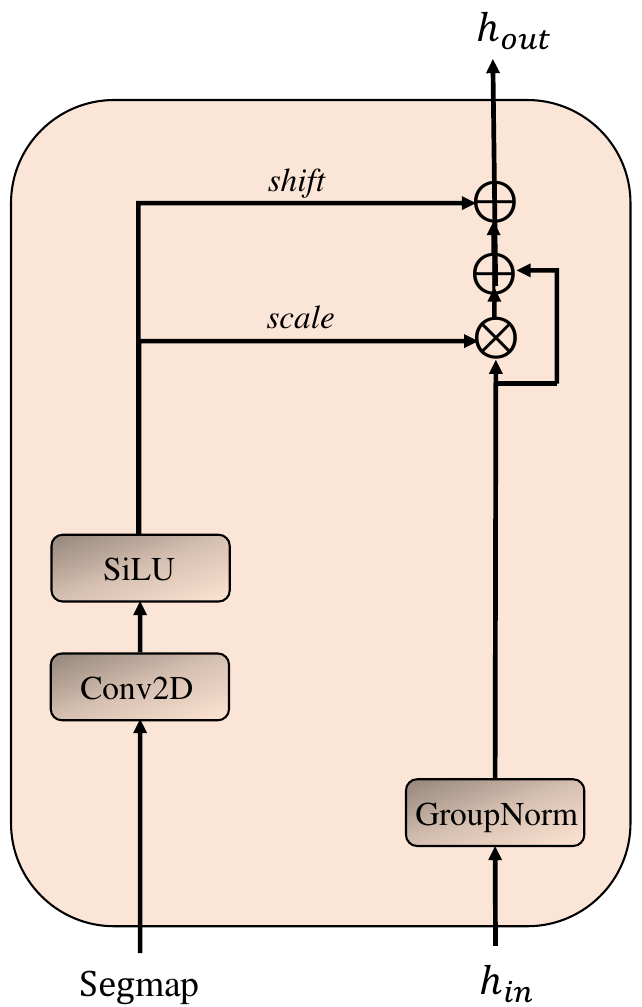}
		\caption{}
		\label{subfig:SPADE}
	\end{subfigure}
	}
	\caption{(a) Details of the SCP ControlNet Encoder Block, highlighting that the final block in the Encoder typically lacks the CrossAttnDownBlock2D layers. (b) Details of SPADE (Spatially-Adaptive Normalization). Here, Segmap represents semantic map features, $h_{in}$ denotes input features, $t_{emb}$ indicates time embedding, and $h_{out}$ signifies output features.}
	\label{fig:SPADE}
\end{figure}

\cref{fig:SCP_ControlNet} provides an overview of the SCP ControlNet.
The adaptations of SCP ControlNet from the vanilla ControlNet focus primarily on two aspects:

\textbf{SCP ControlNet Encoder Block}: As depicted in \cref{subfig:encoder}, the encoder block of SCP ControlNet differs from that of vanilla ControlNet by substituting the GroupNorm in ResNetBlock2D with SPADE.
This modification allows for the preservation of semantic information that would otherwise be "washed away" by GroupNorm and LayerNorm within vanilla ControlNet.
Although the CrossAttnDownBlock2D also contains GroupNorm, we have chosen not to modify it with SPADE in order to preserve the control capability of the text. Additionally, attention layers are typically not set in the last Encoder Block. Hence, the block shown in \cref{subfig:encoder} should be depicted without the CrossAttnDownBlock2D in the last Encoder Block.

\textbf{Multiscale Semantic Information Utilization}: Beyond concatenating semantic information with latent noise, as done in vanilla ControlNet, the presence of SPADE enables the utilization of semantic information across various sizes of encoder blocks.
\cref{subfig:SPADE} illustrates the detailed architecture of SPADE, which requires an input of semantic features. To this end, we designed a Condition Feature Extractor capable of extracting multi-scale features (see \cref{fig:SCP_ControlNet}). 
The use of multi-scale features is common in segmentation tasks\cite{cheng2022masked}, particularly for curvilinear objects, where they significantly enhance model performance. 
We adopted this approach in our generative tasks as well\cite{zhao2024uni}.

Given that SPADE preserves semantic information through normalization, the modified structures do not introduce any additional supervision signals. Consequently, the loss function used to train SCP ControlNet remains consistent with that of vanilla ControlNet, outlined as follows:
\begin{equation}
	\mathcal{L} = \mathbb{E}_{z_0,t,c_t,c_s,\epsilon \sim \mathcal{N}(0,1)} \left[ \left\| \epsilon - \epsilon_{\theta}(z_t, t, c_t, c_s) \right\|_2^2 \right]
\end{equation}
where \(z_0\) represents the input image, \(z_t\) denotes an image that has undergone the diffusion process and is nearly pure noise, following the addition of noise over \(t\) iterations. 
The \(c_t\) stands for the text prompt, and \(c_s\) signifies a special condition, which in this paper refers to the semantic map. 
The function \(\epsilon_\theta\) is a learnable network that predicts the added noise \(\epsilon\). 
\(\mathcal{L}\) constitutes the overall learning objective for the entire diffusion model.

\subsection{Pipeline for expanding datasets}
\begin{figure}[!ht]
	\centering
	\includegraphics[width=0.95\textwidth]{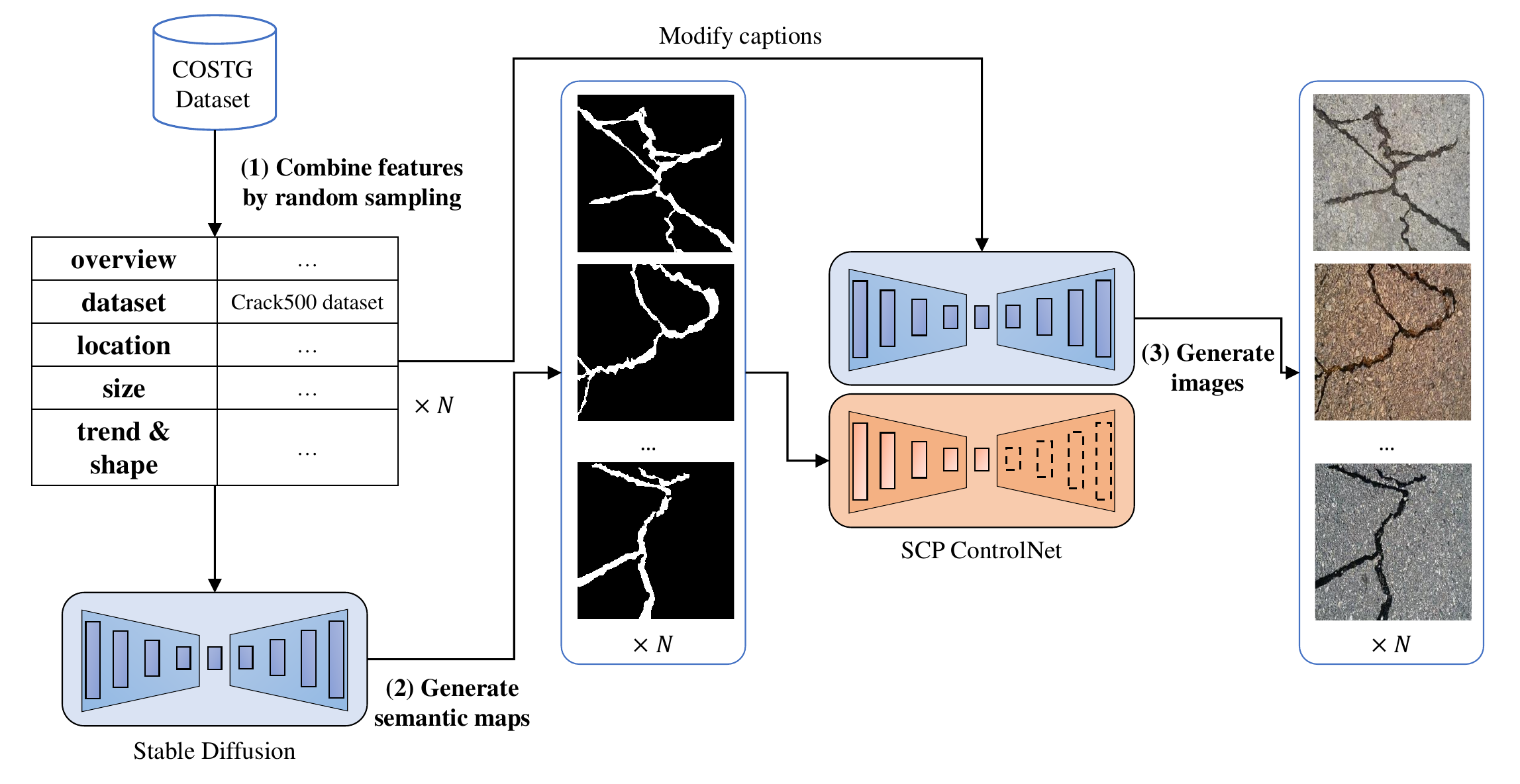}
	\caption{Overview of the dataset expansion pipeline, illustrating the process with the Crack500 dataset example. 
		The pipeline is divided into three steps: (1) combine features through random sampling to obtain captions for generating semantic maps. (2) Utilizing captions for conditional generation to obtain semantic maps. (3) Employing semantic maps and modified captions to generate images.}
	\label{fig:pipeline}
\end{figure}

\cref{fig:pipeline} displays the pipeline for expanding the dataset, with an example on the Crack500 dataset. 
Let \( D_{\text{crack500}}^{\text{seg}} \) be the subset of the COSTG dataset $D_{\text{COSTG}}$ containing semantic maps and their corresponding textual descriptions, and \( D_{\text{crack500}}^{\text{img}} \) be the subset containing images and their image textual descriptions. We define the Crack500 subset of COSTG as \( D_{\text{crack500}} = D_{\text{crack500}}^{\text{seg}} \cup D_{\text{crack500}}^{\text{img}} \subseteq D_{\text{COSTG}} \), where each element in \( D_{\text{crack500}}^{\text{seg}} \) is paired with a corresponding element in \( D_{\text{crack500}}^{\text{img}} \).
The pipeline is divided into three steps:
\begin{itemize}
	\item [1.]\textbf{Combine features through random sampling to obtain captions for generating semantic maps}.

	Sample from \( D_{\text{crack500}} \) to retrieve features \( f_{\text{location}} \), \( f_{\text{size}} \), and \( f_{\text{trend \& shape}} \). Then, sample \( f_{\text{overview}} \) from \( D_{\text{crack500}}^{\text{seg}} \) and the corresponding image overview feature \( f'_{\text{overview}} \) from \( D_{\text{crack500}}^{\text{img}} \) and \( f_{\text{dataset}} = \text{"Crack500 dataset"}\).
	The caption \( C_{\text{sem}} \) for generating a semantic map is then assembled as \( C_{\text{sem}} = f_{\text{overview}} + f_{\text{dataset}} + f_{\text{location}} + f_{\text{size}} + f_{\text{trend \& shape}} \).
	
	\item [2.]\textbf{Utilizing captions for conditional generation to obtain semantic maps}. 
	
	Utilize the caption \( C_{\text{sem}} \) in the conditional generation process to yield a new semantic map \( S' \): \( S' = G(S|C_{\text{sem}}) \), where in this paper $G$ denote Stable Diffusion.
	
	\item [3.]\textbf{Employing semantic maps and modified captions to generate images}. 
	
	The caption \( C_{\text{img}} \) for image generation is composed by updating \( f_{\text{overview}} \) to \( f'_{\text{overview}} \), reflecting the corresponding image’s characteristics, and is given by \( C_{\text{img}} = f'_{\text{overview}} + f_{\text{dataset}} + f_{\text{location}} + f_{\text{size}} + f_{\text{trend \& shape}} + f_{\text{background}} \). The image is then synthesized: \( I' = G'(I|S', C_{\text{img}}) \), where $G'$ in this paper denote the SCP ControlNet.
\end{itemize}
Then we get a pair of data $(I', S')$ for curvilinear object segmentation. 

\section{Experiments}
\subsection{Definition of expanding dataset}
In our experiments, we define the task of expanding dataset as follows:
Let us consider an initial, limited dataset \( D_{\text{orig}} = \{ (x_i, y_i) \}_{i=1}^n \), accompanied by a validation set \( D_{\text{val}} \) that adheres to the same probability distribution, denoted by \( P(D_{\text{orig}}) = P(D_{\text{val}}) \). 
In this notation, \( x_i \) represents the image, and \( y_i \) denotes its corresponding label (semantic map). 
Our goal is to synthesize an expanded dataset \( D_{\text{synth}} = \{ (x'_j, y'_j) \}_{j=1}^m \) where \( m \) is significantly larger than \( n \).

After training a model on the combined dataset \( D_{\text{train}} = D_{\text{orig}} \cup D_{\text{synth}} \), we aim for the performance metric \( M_{\text{train}} \), evaluated on \( D_{\text{val}} \), to exceed the metric \( M_{\text{orig}} \) derived from training exclusively on \( D_{\text{orig}} \). In other words, our target is \( M_{\text{train}} > M_{\text{orig}} \), where an improved metric score \( M \) indicates better performance of the model.

\subsection{Datasets and evaluation}
In the experimental section, we conducted experiments using three types of curvilinear objects (angiography, cracks, and retina) across six public datasets: CHUAC\cite{CHASE_CHUAC2018}, XCAD\cite{XCADma2021self}, DCA1\cite{DCA1cervantes2019automatic}, DRIVE\cite{DRIVEstaal2004ridge}, CHASEDB1\cite{CHASE_CHUAC2018}, and Crack500\cite{Crack500yang2019feature}.

To assess the efficacy of the proposed method for expanding curvilinear object segmentation datasets, we trained vanilla segmentation models (such as UNet) from scratch using the expanded datasets. The models were evaluated using two common metrics for segmentation tasks: mean Intersection over Union (mIoU) and F1 Score\cite{lei2024joint,lei2024integrating,lei2023dynamic}.

In practice, the original dataset was expanding by factors of 2$\times$, 3$\times$, 5$\times$ and so on. 
The training of the segmentation models spanned 100 epochs, with evaluations conducted at the end of each epoch. The best performance was determined using mIoU as the primary benchmark from these evaluations.

\subsection{Comparison methods and settings}
In our comparative experiments, we included established data augmentation techniques such as RandAugment\cite{cubuk2020randaugment} and Cutout\cite{devries2017improved}, alongside GAN-based methods for segmentation dataset expansion. Among these, DPGAN\cite{chai2022synthetic}, specifically designed for medical tumor segmentation dataset expansion, and the Crack Establishment (CE) method\cite{jin2023establishment}, which emphasizes sequential semantic maps and images generation for crack, were noteworthy comparisons. 
Additionally, we evaluated methods based on LDM for semantic image synthesis, including LDM\cite{rombach2022high}, ControlNet\cite{zhang2023adding}, and T2I-adapter\cite{mou2023t2i}.

It is important to note that both GAN-based approaches and LDM for semantic image synthesis follow their generation protocols in their papers without incorporating textual information.
For the experimental settings of ControlNet, T2I-adapter, and the SCP ControlNet proposed in our study, we adhered to the following protocol:
a) Utilize the COSTG dataset (\(D_{\text{COSTG}}\)) to fine-tune the UNet component of Stable Diffusion 1.5 and
b) further fine-tune the additional network (adapter for T2I-adapter and trainable copy of UNet for SCP ControlNet and vanilla ControlNet) using the image subset of COSTG (\(D_{\text{COSTG}}^{\text{img}}\)) and the corresponding semantic maps.

\subsection{Experimental results}
\begin{table}[!ht]
	\centering
	\caption{Performance of vanilla UNet trained from scratch by various expanding methods on various datasets and their 5$\times$ extension.}
	\label{tab:semantic_eval}
	\resizebox{\textwidth}{!}{%
		\begin{tabular}{@{}clcclcccccclcccc@{}}
			\toprule
			\multirow{2}{*}{Dataset} &  & \multicolumn{2}{c}{Crack} &  & \multicolumn{6}{c}{Angiogrophy} &  & \multicolumn{4}{c}{Retina} \\ \cmidrule(lr){3-4} \cmidrule(lr){6-11} \cmidrule(l){13-16} 
			&  & \multicolumn{2}{c}{Crack500} &  & \multicolumn{2}{c}{XCAD} & \multicolumn{2}{c}{DCA1} & \multicolumn{2}{c}{CHUAC} &  & \multicolumn{2}{c}{DRIVE} & \multicolumn{2}{c}{CHASEDB1} \\ \midrule
			Method &  & mIoU & F1 &  & mIoU & F1 & mIoU & F1 & mIoU & F1 &  & mIoU & F1 & mIoU & F1 \\ \midrule
			\textit{Original} &  & 73.2 & 81.4 &  & 81.2 & 88.6 & 63.1 & 77.4 & 51.2 & 67.7 &  & 68.6 & 81.4 & 65.3 & 79.0 \\ \midrule
			\textit{Expanded} &  &  &  &  &  &  &  &  &  &  &  &  &  &  &  \\
			Cutout &  & 74.2 & 81.1 &  & 82.5 & 89.3 & 63.8 & 77.9 & 53.7 & 70.5 &  & 68.3 & 81.1 & 66.9 & 80.2 \\
			RandAugment &  & 74.4 & 81.8 &  & 83.5 & 91.1 & 63.2 & 77.5 & 54.3 & 71.2 &  & 67.2 & 80.4 & 66.2 & 79.6 \\ \midrule
			\textit{GAN-Based} &  &  &  &  &  &  &  &  &  &  &  &  &  &  &  \\
			DPGAN &  & 76.3 & 83.7 &  & 83.7 & 91.4 & 64.5 & 78.1 & 57.3 & 73.3 &  & 70.1 & 81.6 & 67.3 & 80.4 \\
			CE &  & 76.5 & 84.5 &  & 83.9 & 91.6 & 65.5 & 79.2 & 60.0 & 75.2 &  & 70.4 & 82.7 & 68.1 & \textbf{81.5} \\ \midrule
			\textit{LDM-Based} &  &  &  &  &  &  &  &  &  &  &  &  &  &  &  \\
			LDM &  & 76.0 & 83.0 &  & 82.5 & 90.5 & 64.1 & 77.9 & 56.9 & 71.7 &  & 69.7 & 81.0 & 66.7 & 80.0 \\
			ControlNet &  & 75.4 & 84.5 &  & 82.4 & 91.1 & 66.2 & 78.9 & 58.4 & 74.3 &  & 69.3 & 83.1 & 67.1 & 80.4 \\
			T2I-apdater &  & 75.2 & 84.9 &  & 82.2 & 90.9 & 65.3 & 78.4 & 57.9 & 74.0 &  & 68.8 & 80.8 & 65.9 & 79.5 \\ \midrule
			\begin{tabular}[c]{@{}c@{}}SCP ControlNet\\ (ours)\end{tabular} &  & \textbf{78.4} & \textbf{86.3} &  & \textbf{85.3} & \textbf{92.3} & \textbf{68.1} & \textbf{81.2} & \textbf{61.5} & \textbf{76.0} &  & \textbf{71.2} & \textbf{86.9} & \textbf{68.8} & 81.2 \\ \bottomrule
		\end{tabular}%
	}
\end{table}

\begin{figure}[!ht]
	\centering
	\includegraphics[width=0.95\textwidth]{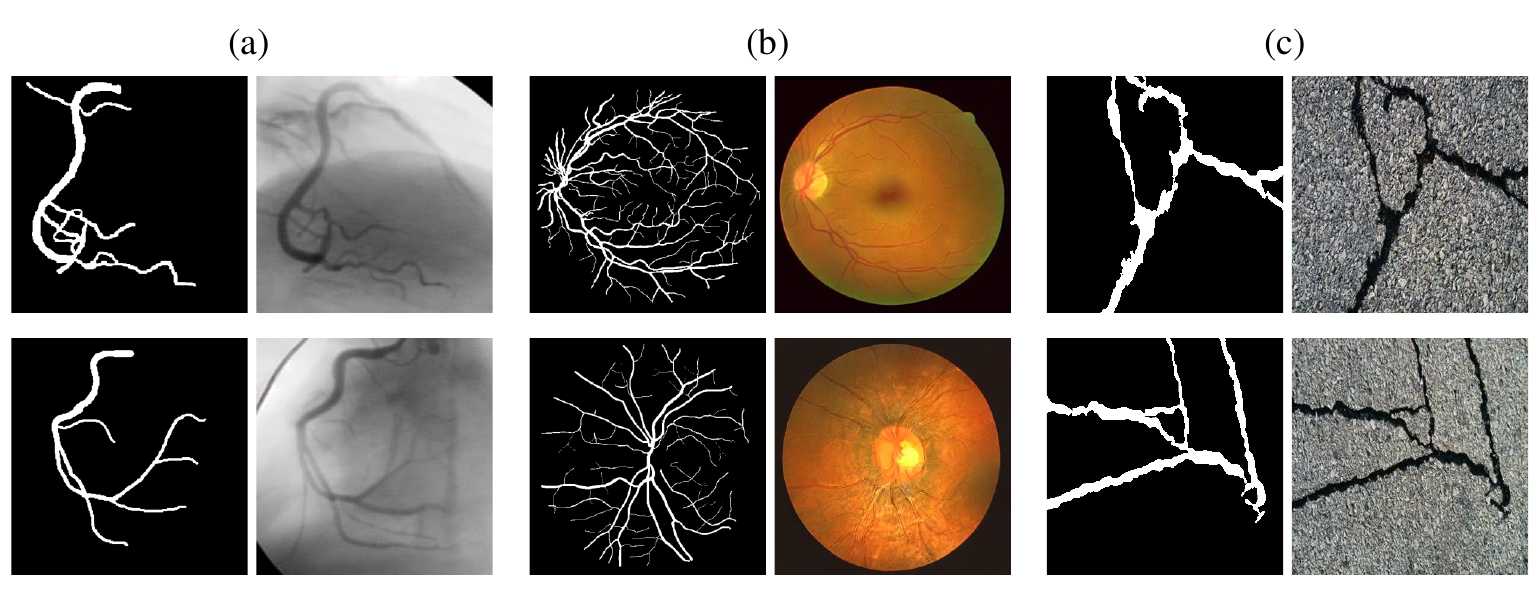}
	\caption{Visual examples of expanded curvilinear segmentation data: (a) angiography, (b) retina, and (c) crack.}
	\label{fig:visual_examples}
\end{figure}

\cref{tab:semantic_eval} showcases the performance of vanilla UNet models trained from scratch across various datasets and their 5$\times$ expansions using different expansion methods. 
Here, "\textit{Original}" denotes the performance after training UNet on the original dataset, while the rest represent performance after training UNet on datasets expanded 5$\times$ using respective methods. 
It's important to note that the generation protocols differ among these methods. Traditional data augmentation techniques, such as Cutout and RandAugment, applied the same transformations to both images and semantic maps to achieve their 5$\times$ dataset expansions, whereas the two GAN-based methods adhered to the generation protocols outlined in their respective papers. Meanwhile, LDM, ControlNet, T2I-adapter, and our proposed SCP ControlNet utilized the COSTG dataset and followed the pipeline proposed in this paper for semantic map generation, subsequently using semantic maps and text for image synthesis (noting that LDM's semantic synthesis does not use text).

In \cref{tab:semantic_eval}, after increasing the dataset size by 5-fold, the effectiveness of our proposed SCP ControlNet is validated, particularly when compared to other LDM-based methods. Further comparison with GAN-based methods and traditional data augmentation techniques underscores the efficacy of our proposed pipeline for expanding segmentation datasets.

\begin{figure}[!ht]
	\centering
	\includegraphics[width=0.95\textwidth]{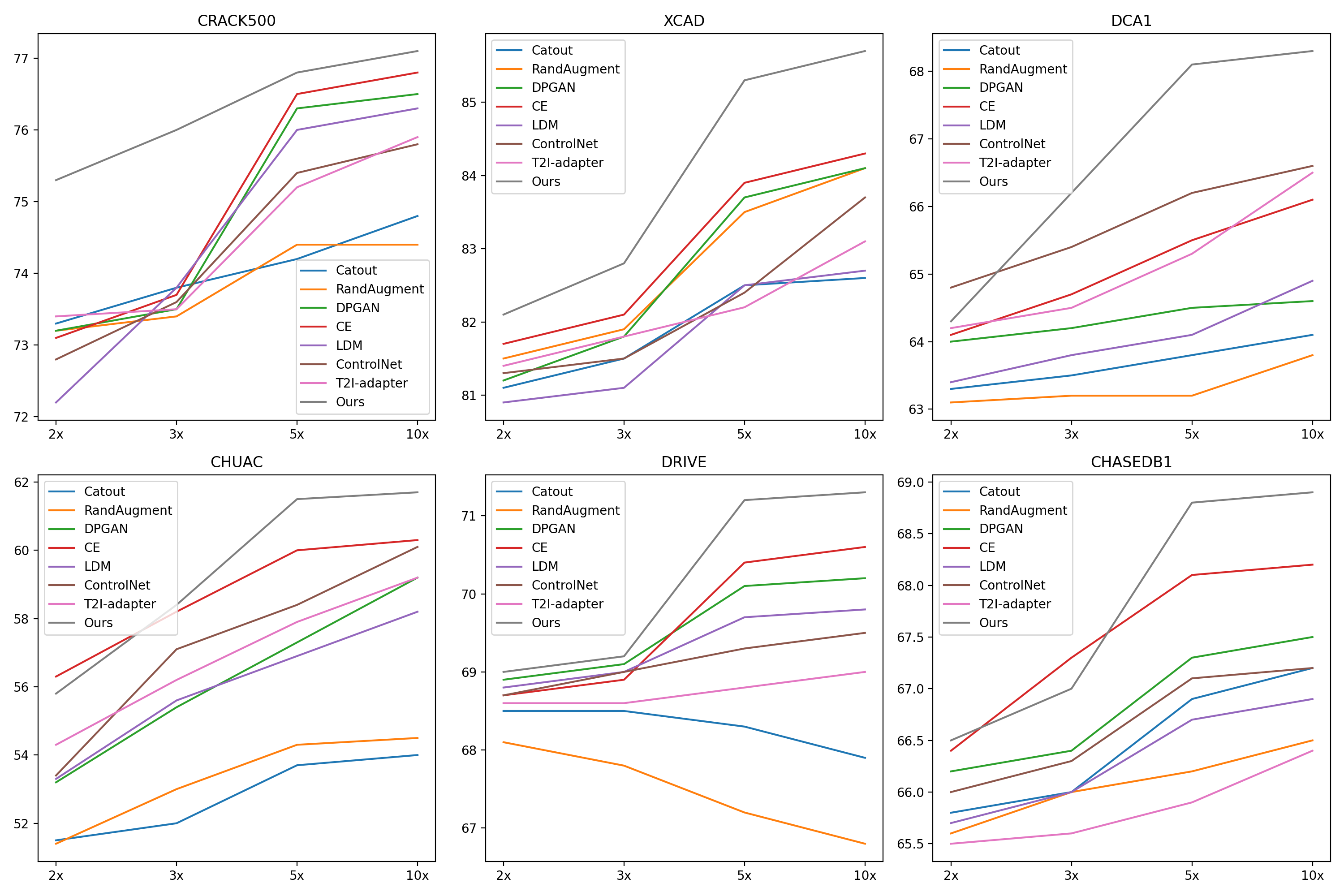}
	\caption{Comparative analysis of UNet's mIoU performance on the six public curvilinear object segmentation datasets across various expansion ratios using different dataset expansion methods.}
	\label{fig:ratio}
\end{figure}

\cref{fig:visual_examples} presents visualizations of samples generated by our proposed pipeline, offering convincing evidence of our proposed SCP ControlNet exceptional capability to produce images of curvilinear objects that closely match the provided semantic maps.

\cref{fig:ratio} shows that our method, with a 5$\times$ dataset expansion, surpasses the performance of alternative methods at a 10$\times$ expansion on some datasets. This superior efficiency suggests that our method's effectiveness is at least twice that of others. Our method not only produces a broader range of samples but also maintains semantic consistency between synthesized images and semantic maps, leading to significantly enhanced model performance.

\section{Conclusion}
In summary, this paper has presented a novel approach to expand datasets for curvilinear object segmentation, significantly enhancing the quality and informativeness of the data. By leveraging textual features to create synthetic images that exceed the original dataset's distribution, we've developed the COSTG dataset, which breaks free from conventional dataset limitations. The Semantic-Consistency-Preserving ControlNet (SCP ControlNet) was introduced, incorporating Spatially-Adaptive Normalization (SPADE) to maintain the semantic consistency of data through the synthesis process. Our extensive experiments across diverse curvilinear objects and datasets affirm the superior performance of our approach.

\newpage
\section*{Acknowledgements}
The authors would like to extend their sincere gratitude to the following funding sources for their support: the National Natural Science Foundation of China (Grant No. 62176029), and the Chongqing Municipal Technology Innovation and Application Development Special Program (Grant Nos. CSTB2023TIAD-KPX0064, CSTB2022TIAD-KPX0206).

\bibliographystyle{splncs04}
\bibliography{main}

\newpage
\section{Appendix}
The appendix is structured into three principal sections. \cref{sec:A1} focuses on the COSTG dataset, featuring additional visualization examples (\cref{sec:A1.1}), prompts for generating annotations (\cref{sec:A1.2}), and insights into the application of Language-Vision Models (LVMs) (\cref{sec:A1.3} and \cref{sec:A1.4}). \cref{sec:A2} contains supplementary notes on various experiments, complemented by additional visualization examples.  \cref{sec:A3} offers prompts related to the images presented in the paper.

A DEMO and related code and the COSTG dataset will be made public.\footnote{\url{https://github.com/tanlei0/COSTG}}

\subsection{More details about COSTG}
\label{sec:A1}
\subsubsection{Some examples about COSTG}
\label{sec:A1.1}
\begin{figure}[!ht]
	\centering
	\includegraphics[width=\textwidth]{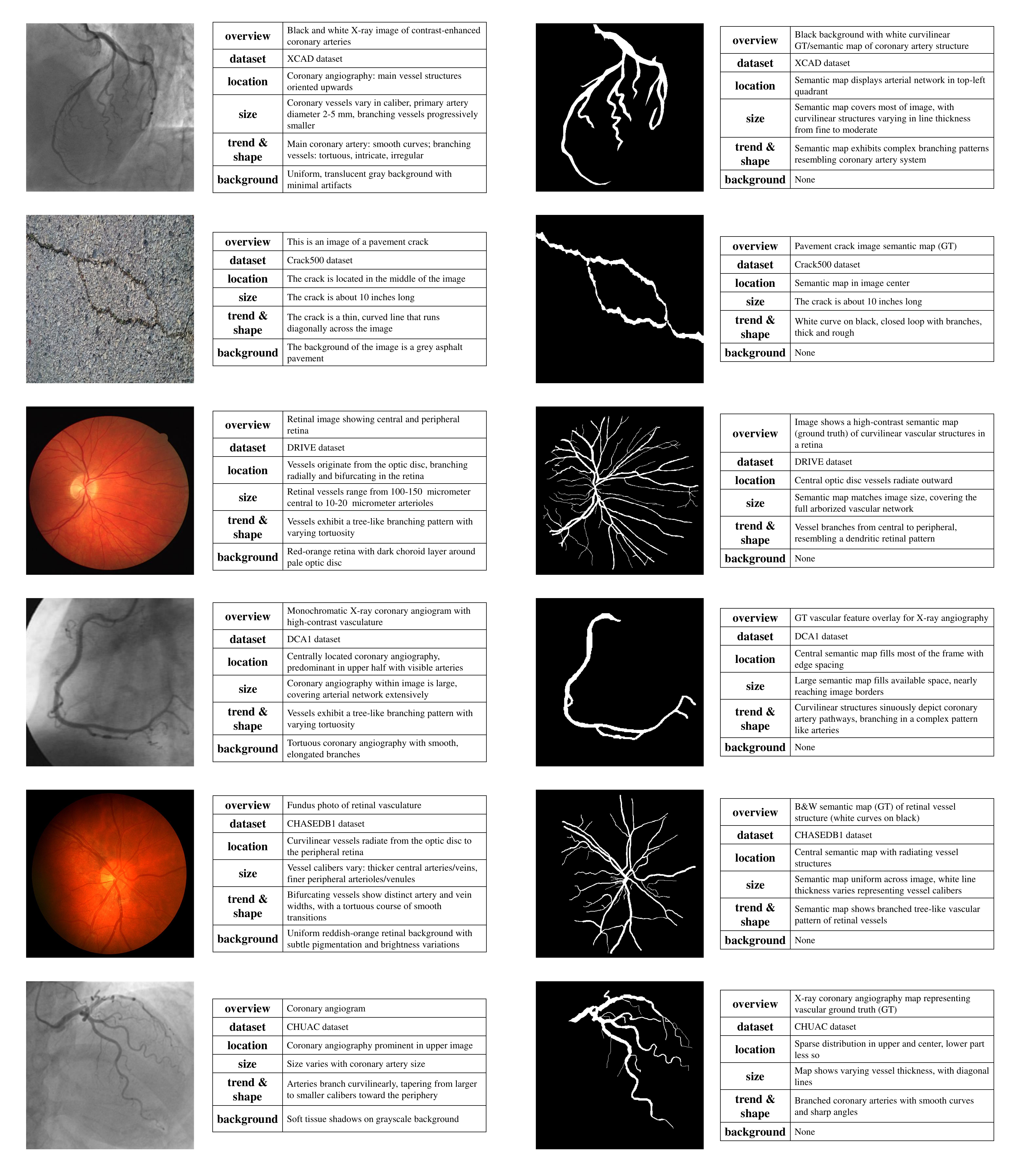}
	\caption{Examples from the COSTG Dataset. On the left: Images alongside their textual descriptions. On the right: Corresponding semantic maps with textual descriptions. Displayed in sequence from top to bottom, these samples originate from six public datasets: XCAD\cite{XCADma2021self}, Crack500\cite{Crack500yang2019feature}, DRIVE\cite{DRIVEstaal2004ridge}, DCA1\cite{DCA1cervantes2019automatic}, CHASEDB1\cite{CHASE_CHUAC2018}, and CHUAC\cite{CHASE_CHUAC2018}.}
	\label{fig:6examples}
\end{figure}

\cref{fig:6examples} showcases samples created based on these public datasets. 
On the left, images are paired with their textual descriptions, and on the right are the corresponding semantic maps with textual descriptions. Displayed from top to bottom, the samples originate from the six public datasets: XCAD\cite{XCADma2021self}, Crack500\cite{Crack500yang2019feature}, DRIVE\cite{DRIVEstaal2004ridge}, DCA1\cite{DCA1cervantes2019automatic}, CHASEDB1\cite{CHASE_CHUAC2018} and CHUAC\cite{CHASE_CHUAC2018}

\subsubsection{Prompts for generating annotations}
\label{sec:A1.2}
Both ChatGPT-4-V and Gemini 1.5 were finely tuned with instructions and possess the capability to identify images of curvilinear objects. Therefore, we crafted specific prompts for different datasets to facilitate the annotation of COSTG.
 
For \textbf{angiography data}, we designed the following two types of prompts for images and semantic maps, respectively:
\begin{itemize}
	\item For images: 

	"Now, you are a research assistant helping me to capture some features of some curvilinear objects in an image in a literal way. Please note that this conversation is purely about analyzing images to help me achieve a literal capture of features of curvilinear objects in a large model-based research project.
	This is an X-ray coronary angiography image from the \{dataset\} dataset. Please generate a caption for this image, which along with this image will be used to train Stable Diffusion. Please note that the generated caption must be concise and informative and should contain the following information separated by semicolons: 
	\begin{itemize}
		\item[(1)] a concise description of the entire image; 
		\item[(2)] the name of the dataset; 
		\item[(3)] the location of the coronary angiography within the image; 
		\item[(4)] the size of the coronary angiography; 
		\item[(5)] the trend and shape of the coronary angiography; 
		\item[(6)] a brief description of the background.
	\end{itemize}
	Note that your response should contain only the above 6 points and indicate them with serial numbers 1-6. In order for the generated caption to be informative enough for training, 3-5 should be very detailed and contain as much detail as possible and 6 can be very concise."
	\item For semantic maps: 
	
	"Now, you are a research assistant helping me to capture some features of some curvilinear objects in an image in a literal way. Please note that this conversation is purely about analyzing images to help me achieve a literal capture of features of curvilinear objects in a large model-based research project.
	This is a semantic map (ground true (GT)) of X-ray coronary angiography image from the \{dataset\} dataset. 
	Generate a caption for this semantic map, and this caption and this image will be used to train Stable Diffusion. Note that the generated caption needs to be concise and informative, and should contain the following information separated by semicolons: 
	\begin{itemize}
		\item[(1)] a concise description of the entire image, clearly stating that it is a semantic map or ground truth (GT);
		\item[(2)] the name of the dataset;
		\item[(3)] the location of the semantic map (curvilinear objects) within the image; 
		\item[(4)] the size of the semantic map (curvilinear objects); 
		\item[(5)] the trend and shape of the semantic map (curvilinear objects).
	\end{itemize}
	Note that your response should contain only the above 5 points and indicate them with serial numbers 1-5. In order for the generated caption to be informative enough for training, 3-5 should be very detailed and contain as much detail as possible."
\end{itemize}

For \textbf{retina data}, we designed the following two types of prompts for images and semantic maps, respectively:
\begin{itemize}
	\item For images: 
	
	"Now, you are a research assistant helping me to capture some features of some curvilinear objects in an image in a literal way. Please note that this conversation is purely about analyzing images to help me achieve a literal capture of features of curvilinear objects in a large model-based research project.
	This is a retinal image from the \{dataset\} dataset. Please generate a caption for this image, which along with this image will be used to train stable diffusion. Please note that the generated caption must be concise and informative and should contain the following information separated by semicolons: 
	\begin{itemize}
		\item[(1)] a concise description of the entire image; 
		\item[(2)] the name of the dataset; 
		\item[(3)] the location of the curvilinear objects (vessels) in retinal image; 
		\item[(4)] the size of the curvilinear objects (vessels) in retinal image; 
		\item[(5)] the trend and shape of the coronary angiography; 
		\item[(6)] a brief description of the background.
	\end{itemize}
	Note that your response should contain only the above 6 points and indicate them with serial numbers 1-6. In order for the generated caption to be informative enough for training, 3-5 should be very detailed and contain as much detail as possible and 6 can be very concise."
	\item For semantic maps: 
	
	"Now, you are a research assistant helping me to capture some features of some curvilinear objects in an image in a literal way. Please note that this conversation is purely about analyzing images to help me achieve a literal capture of features of curvilinear objects in a large model-based research project.
	This is a semantic map (groundtrue (GT)) of of the curvilinear objects (vessels) in the retinal image from the \{dataset\} dataset. Generate a caption for this semantic map, and this caption and this image will be used to train stable diffusion. Note that the generated caption needs to be concise and informative, and should contain the following information separated by semicolons: 
	\begin{itemize}
		\item[(1)] a concise description of the entire image, making sure to point out that this image is a semantic map or groundtrue (GT);
		\item[(2)] the name of the dataset;
		\item[(3)] the location of the semantic map (curvilinear objects) within the image; 
		\item[(4)] the size of the semantic map (curvilinear objects); 
		\item[(5)] the trend and shape of the semantic map (curvilinear objects).
	\end{itemize}
	Note that your response should contain only the above 5 points and indicate them with serial numbers 1-5. In order for the generated caption to be informative enough for training, 3-5 should be very detailed and contain as much detail as possible."
\end{itemize}

For \textbf{crack data}, we designed the following two types of prompts for images and semantic maps, respectively:
\begin{itemize}
	\item For images: 
	
	"Now, you are a research assistant helping me to capture some features of some curvilinear objects in an image in a literal way. Please note that this conversation is purely about analyzing images to help me achieve a literal capture of features of curvilinear objects in a large model-based research project.
	This is a pavement crack image from the \{dataset\} dataset. Please generate a caption for this image, which along with this image will be used to train stabilization diffusion. Please note that the generated caption must be concise and informative and should contain the following information separated by semicolons: 
	\begin{itemize}
		\item[(1)] a concise description of the entire image; 
		\item[(2)] the name of the dataset; 
		\item[(3)] the location of the crack in the image; 
		\item[(4)] the size of the crack; 
		\item[(5)] trend and shape of the crack; 
		\item[(6)] a brief description of the background.
	\end{itemize}
	Note that your response should contain only the above 6 points and indicate them with serial numbers 1-6. In order for the generated caption to be informative enough for training, 3-5 should be very detailed and contain as much detail as possible and 6 can be very concise."
	\item For semantic maps: 
	
	"Now, you are a research assistant helping me to capture some features of some curvilinear objects in an image in a literal way. Please note that this conversation is purely about analyzing images to help me achieve a literal capture of features of curvilinear objects in a large model-based research project.
	This is a semantic map (groundtrue (GT)) of pavement crack image from the \{dataset\} dataset. Generate a caption for this semantic map, and this caption and this image will be used to train stable diffusion. Note that the generated caption needs to be concise and informative, and should contain the following information separated by semicolons: 
	\begin{itemize}
		\item[(1)] a concise description of the entire image, making sure to point out that this image is a semantic map or groundtrue (GT);;
		\item[(2)] the name of the dataset;
		\item[(3)] the location of the semantic map (curvilinear objects, cracks) within the image; 
		\item[(4)] the size of the semantic map (curvilinear objects, cracks); 
		\item[(5)] the trend and shape of the semantic map (curvilinear objects, cracks).
	\end{itemize}
	Note that your response should contain only the above 5 points and indicate them with serial numbers 1-5. In order for the generated caption to be informative enough for training, 3-5 should be very detailed and contain as much detail as possible."
\end{itemize}

where \{dataset\} should be replaced the name of dataset. This structured prompts ensure that the generated captions are sufficiently informative for training, with a detailed emphasis on the critical features of the curvilinear objects.

\subsubsection{About the use of ChatGPT-4-V and Gemini 1.5}
\label{sec:A1.3}
\cref{fig:LVM} presents annotation examples using ChatGPT-4-V and Gemini 1.5.
It is observed that ChatGPT-4-V tends to produce more verbose and redundant output, and notably, it does not specify the length and width of curvilinear objects. 
In contrast, Gemini 1.5 generates outputs that are more concise and includes the length and width of curvilinear targets. However, this does not imply that Gemini 1.5 surpasses ChatGPT-4-V in annotating visual features of curvilinear objects. 
On the contrary, Gemini 1.5 shows inferior performance in understanding instructions and is more prone to errors compared to ChatGPT-4-V. 
Due to ChatGPT-4-V's stricter usage fees and token limitations compared to Gemini 1.5, for the COSTG dataset, approximately 700 entries for angiography and retina categories, including datasets XCAD, CHUAC, DCA1, DRIVE, and CHASEDB1, were annotated using ChatGPT-4-V (It cost roughly 10 dollars to build that part of the COSTG dataset). 
Meanwhile, approximately 700 entries for the crack category, including the Crack500 dataset, were annotated using Gemini 1.5.

Additionally, Gemini 1.5 was employed to condense the annotations made by ChatGPT-4-V. 
This was necessitated by the limit of number of  token 77 in SD 1.5's CLIP, whereas the text of curvilinear object features generated by ChatGPT-4-V significantly exceeded this limitation. Consequently, in the fine-tuning of SD 1.5, we had to abbreviate the captions and omit the "background" field to comply with the token restrictions.

During our exploration of ChatGPT-4-V and Gemini 1.5, we encountered some issues:
\begin{itemize}
	\item [(1)] Misinterpretation of instructions: For instance, when requesting the size of a curvilinear target, both models might respond with the overall dimensions of the semantic map (e.g., 512x512 pixels), with Gemini 1.5 exhibiting this behavior more frequently (noting that both contributed to approximately 700 annotations each).
	\item [(2)] Combined responses: Both models occasionally merged details about the "size" and "trend \& shape" into a single response, occurring at similar rates for both.
	\item [(3)] Lack of precision in responses: As acknowledged in their official documentation, neither model is capable of accurately recognizing image shapes, reflecting a limitation in the current generation of Language-Vision Models (LVMs).
\end{itemize}

\subsubsection{A note about other LVMs in building COSTG}
\label{sec:A1.4}
\begin{figure}[!ht]
	\centering
	\includegraphics[width=0.9\textwidth]{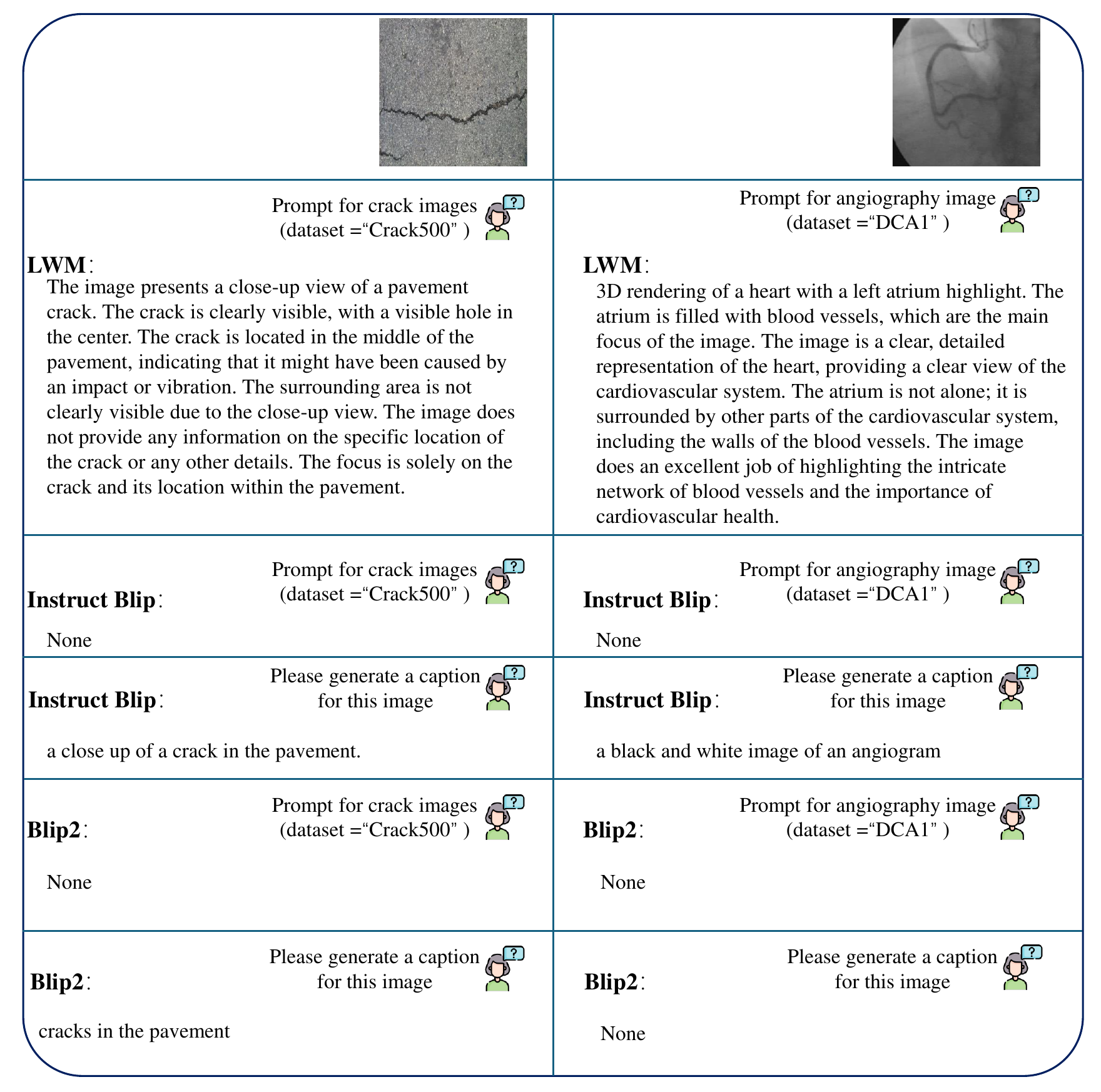}
	\caption{Other LVMs in building COSTG}
	\label{fig:oLVM}
\end{figure}

In the initial phases of our experiments, besides ChatGPT-4-V and Gemini 1.5, we also explored several other relatively recent Language-Vision Models (LVMs) for our annotation tasks, including BLIP2\cite{li2023blip2}, Instruct BLIP\cite{instructblip}, and LWM\cite{liu2023world}. Despite the outstanding performance reported in their respective publications, their generalization capabilities, particularly within the domain of curvilinear objects, raised concerns, as highlighted by the "no free lunch" theorem. \cref{fig:oLVM} provides illustrative examples of these limitations. 
It is observed that LWM struggles with accurately interpreting instructions and exhibits limited generalization capabilities. In contrast, BLIP2 and Instruct BLIP are notably challenged by complex, lengthy instructions, managing only to comprehend simpler directives.

\subsection{More details about the experiments}
\label{sec:A2}
\subsubsection{Datasets}
For the six public datasets referenced in this paper, we adhere to the training and testing set splits specified in their respective publications, with the exception of XCAD. 
Due to the absence of semantic maps in XCAD's training set, we redefined its training and testing divisions. 
Detailed information on this partitioning is available in the open-source dataset.
Moreover, to streamline the training process for the generative models, we consistently resized both images and semantic maps to 512$\times$512 pixels.

\subsubsection{Protocols for experiments}

It is important to note that the two GAN-based methods and the LDM approach for semantic image synthesis each adhere to distinct generation protocols and do not incorporate textual information.
For the settings of ControlNet, T2i-adapter, and the PSC ControlNet proposed in our study, we implemented the following configurations:
1. Utilized the COSTG dataset (\(D_{\text{COSTG}}\)) to fine-tune the UNet architecture of Stable Diffusion.
2. Conducted additional fine-tuning of networks using the \(D_{\text{COSTG}}^{\text{img}}\) and corresponding semantic maps (segmap).
3. Determined the selection index for the weight of Stable Diffusion and additional networks based on human evaluation to identify the optimal one among the 10 lowest FID values. This step is critical because the fine-tuning process is highly susceptible to overfitting, typically occurring after 6000 steps when the batch size is set to 16.

\subsubsection{More visualization}
\begin{figure}[!ht]
	\centering
	\includegraphics[width=0.9\textwidth]{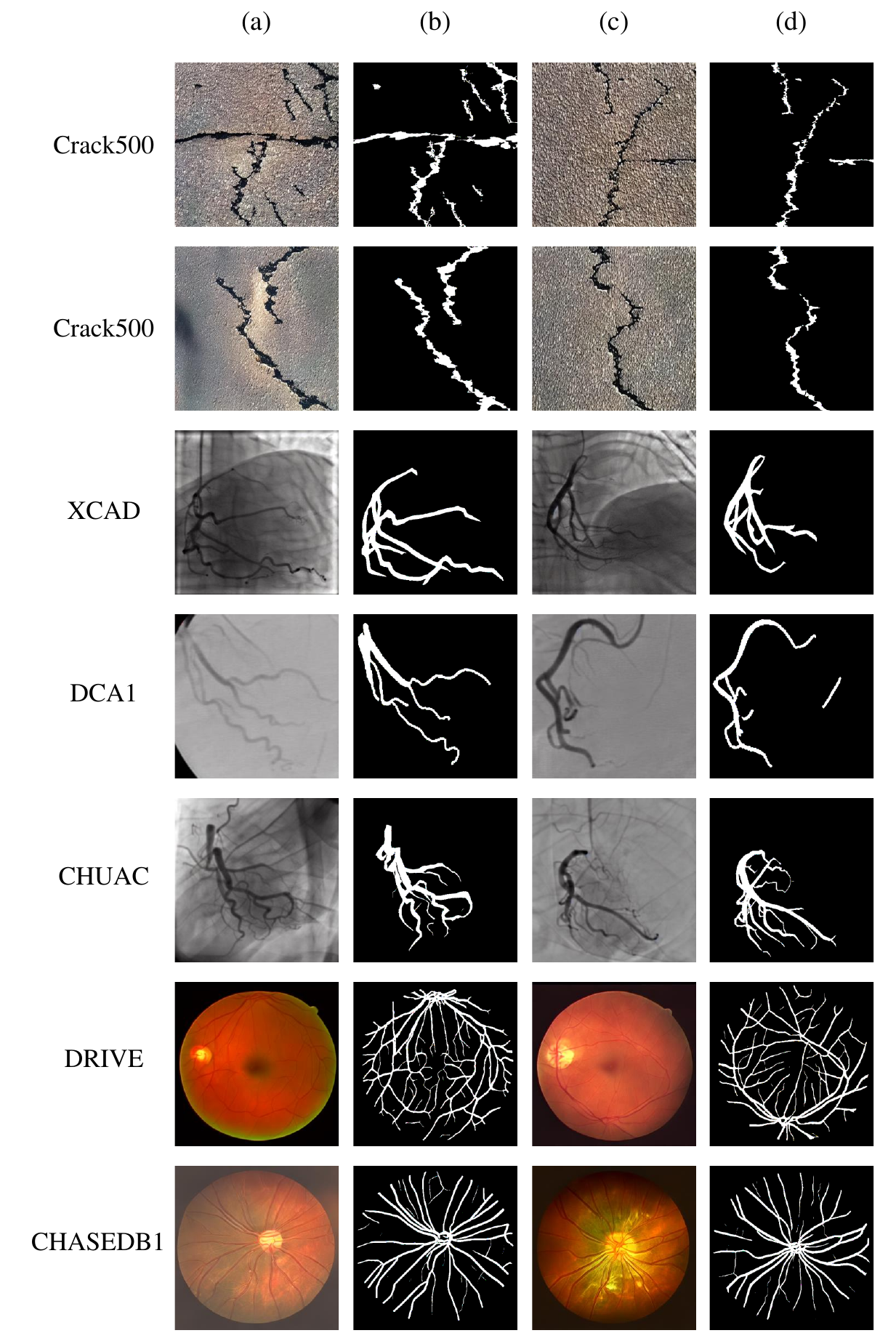}
	\caption{More visualization}
	\label{fig:more_examples}
\end{figure}

\cref{fig:more_examples} presents additional visualization examples, with their corresponding prompts detailed in \cref{sec:more_examples_prompts}.

\subsubsection{Tricks for generating reasonable semantic maps}
In generating semantic maps, we employed some tricks  to ensure the validity of the outputs.
\begin{itemize}
	\item [1.] \textbf{Textual description sampling is restricted to within-dataset sources}. For instance, when expanding the DCA1 dataset using the COSTG dataset—which also includes XCAD and CHUAC—the textual descriptions are sourced exclusively from DCA1. This practice is essential to maintain the coherence of generated semantic maps and avoid errors from cross-dataset mixing.
	\item [2.] \textbf{The generated semantic maps are binarized using the Otsu thresholding method}. 
\end{itemize}

\subsubsection{Ablation study for SCP ControlNet}:
\begin{table}
	\centering
	\caption{Ablation study of SCP ControlNet on 5$\times$ expanded Crack500 dataset}
	\label{tab:ablation}
	
	\resizebox{.8\textwidth}{!}{
		\begin{tabular}{@{}ccccccc@{}}
			\toprule
			Module                          & Down.1      & Down.2      & Down.3      & Down.4      & Middle      & mIoU \\ \midrule
			Vanilla ControlNet              &             &             &             &             &             & 75.4 \\ \midrule
			\multirow{5}{*}{SCP ControlNet} & \checkmark  &             &             &             &             & 75.7 \\
			& \checkmark  & \checkmark  &             &             &             & 75.9 \\
			& \checkmark  & \checkmark  & \checkmark  &             &             & 77.9 \\
			& \checkmark  & \checkmark  & \checkmark  & \checkmark  &             & 78.2 \\
			& \checkmark  & \checkmark  & \checkmark  & \checkmark  & \checkmark  & 78.4 \\ \midrule
			\multirow{5}{*}{SCP ControlNet} & \checkmark  &             &             &             &             & 75.7 \\
			&             & \checkmark  &             &             &             & 75.5 \\
			&             &             & \checkmark  &             &             & 77.6 \\
			&             &             &             & \checkmark  &             & 77.3 \\
			&             &             &             &             & \checkmark  & 75.7 \\ \bottomrule
		\end{tabular}%
	}
\end{table}
As shown in Table \ref{tab:ablation}, we have executed an ablation study to explore the impact of integrating SPADE at various stages of our SCP ControlNet. In this table, 'Down.x' and 'Middle' indicate whether SPADE was implemented in the x-th Encoder Block and the Middle Block, respectively, in conjunction with the corresponding size feature from the Condition Feature Extractor. 
The "Vanilla ControlNet" configuration does not include any SPADE adaptations and only merges the final output of the Condition Feature Extractor with noise.
Experimental results indicate that the inclusion of SPADE adaptations in the "Down.3" and "Down.4" stages significantly enhances the network's ability to maintain semantic consistency between the generated images and the conditional semantic maps. 

\subsubsection{Experiments on more segmentation architectures}
\begin{table}[!ht]
	\centering
	\caption{Performance (mIoU) of various segmentation models trained from scratch by various expanding methods on various \textit{angiography} datasets and their 5$\times$ extension.}
	\label{tab:architectures_angiography}
	\resizebox{\columnwidth}{!}{%
		\begin{tabular}{@{}ccccccccccccc@{}}
			\toprule
			\multirow{2}{*}{}                                               &  & \multicolumn{3}{c}{XCAD}                      &  & \multicolumn{3}{c}{DCA1}                      &  & \multicolumn{3}{c}{CHUAC}                     \\ \cmidrule(lr){3-5} \cmidrule(lr){7-9} \cmidrule(l){11-13} 
			&  & UNet          & DeepLabV3+    & Mask2Former   &  & UNet          & DeepLabV3+    & Mask2Former   &  & UNet          & DeepLabV3+    & Mask2Former   \\ \midrule
			\textit{Original}                                               &  & 81.2          & 81.1          & 80.8          &  & 63.1          & 63.4          & 66.1          &  & 51.2          & 52.1          & 50.1          \\ \midrule
			\textit{Expanded}                                               &  &               &               &               &  &               &               &               &  &               &               &               \\
			Cutout                                                          &  & 82.5          & 82.3          & 81.5          &  & 63.8          & 64.1          & 67.3          &  & 53.7          & 54.1          & 51.2          \\
			RandAugment                                                     &  & 83.5          & 83.0          & 82.3          &  & 63.2          & 64.0          & 67.1          &  & 54.3          & 54.5          & 50.9          \\ \midrule
			\textit{GAN-Based}                                              &  &               &               &               &  &               &               &               &  &               &               &               \\
			DPGAN                                                           &  & 83.7          & 84.6          & 82.8          &  & 64.5          & 65.2          & 68.4          &  & 57.3          & 57.5          & 55.3          \\
			CE                                                              &  & 83.9          & \textbf{85.7}          & 83.4          &  & 65.5          & 66.1          & \textbf{69.0} &  & 60.0          & 59.3          & 56.4          \\ \midrule
			\textit{LDM-Based}                                              &  &               &               &               &  &               &               &               &  &               &               &               \\
			LDM                                                             &  & 82.5          & 83.4          & \textbf{82.4} &  & 64.1          & 65.3          & 68.1          &  & 56.9          & 57.2          & 54.1          \\
			ControlNet                                                      &  & 82.4          & 83.3          & 81.9          &  & 66.2          & 66.7          & 68.7          &  & 58.4          & 58.8          & 54.6          \\
			T2I-apdater                                                     &  & 82.2          & 82.9          & 82.1          &  & 65.3          & 65.9          & 68.3          &  & 57.9          & 58.1          & 54.2          \\ \midrule
			\begin{tabular}[c]{@{}c@{}}PSC ControlNet\\ (ours)\end{tabular} &  & \textbf{85.3} & 84.3 & 82.3          &  & \textbf{68.1} & \textbf{67.0} & 68.9          &  & \textbf{61.5} & \textbf{60.9} & \textbf{57.1} \\ \bottomrule
		\end{tabular}%
	}
\end{table}

\begin{table}[!ht]
	\centering
	\caption{Performance (mIoU) of various segmentation models trained from scratch by various expanding methods on various \textit{retina} datasets and their 5$\times$ extension.}
	\label{tab:architectures_retina}
	\resizebox{0.8\columnwidth}{!}{%
		\begin{tabular}{@{}cccclccc@{}}
			\toprule
			\multirow{2}{*}{}                                               & \multicolumn{3}{c}{DRIVE}                     &  & \multicolumn{3}{c}{CHASEDB1}                  \\ \cmidrule(lr){2-4} \cmidrule(l){6-8} 
			& UNet          & DeepLabV3+    & Mask2Former   &  & UNet          & DeepLabV3+    & Mask2Former   \\ \midrule
			\textit{Original}                                               & 68.6          & 68.8          & 66.3          &  & 65.3          & 66.1          & 68.4          \\ \midrule
			\textit{Expanded}                                               &               &               &               &  &               &               &               \\
			Cutout                                                          & 68.3          & 69.1          & 65.4          &  & 66.9          & 67.6          & 70.1          \\
			RandAugment                                                     & 67.2          & 68.3          & 65.1          &  & 66.2          & 67.1          & 70.4          \\ \midrule
			\textit{GAN-Based}                                              &               &               &               &  &               &               &               \\
			DPGAN                                                           & 70.1          & \textbf{71.6}          & 66.4          &  & 67.3          & 68.0          & 70.9          \\
			CE                                                              & 70.4          & 71.4          & 66.8          &  & 68.1          & \textbf{69.3} & 71.2 \\ \midrule
			\textit{LDM-Based}                                              &               &               &               &  &               &               &               \\
			LDM                                                             & 69.7          & 70.4          & 66.1 &  & 66.7          & 67.5          & 69.6          \\
			ControlNet                                                      & 69.3          & 70.2          & 65.8          &  & 67.1          & 67.8          & 69.8          \\
			T2I-apdater                                                     & 68.8          & 69.7          & 65.3          &  & 65.9          & 67.1          & 69.2          \\ \midrule
			\begin{tabular}[c]{@{}c@{}}PSC ControlNet\\ (ours)\end{tabular} & \textbf{71.2} & 70.8 & \textbf{67.1} &  & \textbf{68.8} & 69.0 & \textbf{71.5} \\ \bottomrule
		\end{tabular}%
	}
\end{table}

\begin{table}[!ht]
	\centering
	\caption{Performance (mIoU) of various segmentation models trained from scratch by various expanding methods on \textit{Crack500} dataset and their 5$\times$ extension.}
	\label{tab:architectures_crack}
	\resizebox{0.8\columnwidth}{!}{%
		\begin{tabular}{@{}cccc@{}}
			\toprule
			& UNet          & DeepLabV3+    & Mask2Former   \\ \midrule
			\textit{Original}                                               & 73.2          & 73.3          & 74.9          \\ \midrule
			\textit{Expanded}                                               &               &               &               \\
			Cutout                                                          & 74.2          & 73.7          & 74.8          \\
			RandAugment                                                     & 74.4          & 73.5          & 75.1          \\ \midrule
			\textit{GAN-Based}                                              &               &               &               \\
			DPGAN                                                           & 76.3          & 75.7          & 76.8          \\
			CE                                                              & 76.5          & 75.8          & 77.2          \\ \midrule
			\textit{LDM-Based}                                              &               &               &               \\
			LDM                                                             & 76.0          & 76.2          & 78.6          \\
			ControlNet                                                      & 75.4          & 76.8          & 79.1          \\
			T2I-apdater                                                     & 75.2          & 75.9          & 77.9          \\ \midrule
			\begin{tabular}[c]{@{}c@{}}PSC ControlNet\\ (ours)\end{tabular} & \textbf{78.4} & \textbf{78.1} & \textbf{80.1} \\ \bottomrule
		\end{tabular}%
	}
\end{table}

\cref{tab:architectures_angiography}, \cref{tab:architectures_retina}, and \cref{tab:architectures_crack} showcase the experimental results (mIoU as the metric) of various segmentation architectures across six public datasets, including vanilla UNet\cite{ronneberger2015u}, DeepLabV3+\cite{chen2018encoder}, and Mask2Former\cite{cheng2022masked}. 
UNet and DeepLabV3+ represent convolutional architectures within segmentation networks, while Mask2Former is a segmentation architecture akin to DETR\cite{carion2020end}, based on transformers (note: this does not imply that the encoder or backbone is a transformer). These tables reveal that our method for dataset expansion significantly enhances the performance of convolutional architectures. Due to the inherent design of Mask2Former, its performance on certain curvilinear object datasets, such as DRIVE and CHUAC, is not optimal (reasons for which merit further research). However, our proposed method has already been able to improve the performance of Mask2Former.

\subsection{Prompts for images in the paper}
\label{sec:A3}
\subsubsection{Prompts for \cref{fig:compare_1}}
\begin{itemize}
	\item [(a)] This is an image of a crack in asphalt pavement; Crack500 dataset; The crack is located in the middle of the image; The crack is about 1.5 inches wide and 8 inches long; The crack is jagged and has a Y shape; The background is grey asphalt pavement;
	\item [(b)] Coronary angiogram shows contrast-filled arteries; DCA1 dataset; Central angiography shows heart vessels; Vessel diameters range from small to medium; Angiography shows normal coronary anatomy with branching, smooth curves, and tapering;
	\item [(c)] Fundus photo of eye interior (retina, optic disc, macula, blood vessels); DRIVE dataset; Curvilinear vessels radiate from the optic disc, concentrating at the center and fanning outward; Vessel size varies, larger at optic disc, finer towards periphery; Branched, curvilinear vessels diminish with bifurcation, mostly linear with retinal contour curvatures;
\end{itemize}
\subsubsection{Prompts for \cref{fig:visual_examples}}
\begin{itemize}
	\item [(a)] 
	 \textbf{Top}:
	 Coronary angiogram X-ray shows blood vessel opacification; DCA1 dataset; Central coronary artery branching from aorta; Vessel size variable, proximal segments wider than distal; Vessels are mostly curved, tortuous, smooth, and bifurcated;
	 
	 \textbf{Bottom}:
	 Coronary arteries enhanced in X-ray image; DCA1 dataset; Angiography shows contrast flow in coronary arteries; Coronary angiography reveals normal caliber vessels; Arteries show a smooth, sinuous course with no stenosis or plaque;
	\item [(b)] 
	\textbf{Top}:
	High-res fundus photo of human retina; DRIVE dataset; Optic disc vessels branch radially, densest near the disc; Retinal vessels range from thick central arteries/veins to fine peripheral capillaries; Arboreal vessels transition from straight to tortuous, forming a fractal network;
	
	\textbf{Bottom}:
	Color fundus photograph of human retina, showing optic disc, macula, vessels, and background (CHASEDB1). Curvilinear vessels distributed throughout; CHASEDB1 dataset; Main vessels thicken near optic disc, then taper with branching; Finer vessels branch and curve away from the optic disc; Arterioles: light, narrow; Venules: dark, wide;
	
	\item [(c)] 
	\textbf{Top}:
	This is an image of a crack in asphalt pavement; Crack500 dataset; The crack is located in the middle of the image; The crack is about 1.5 inches wide; The crack is linear and has a slight curve to the right; The background is a close-up of asphalt pavement with small pebbles;
	
	\textbf{Bottom}
	This is an image of a crack in asphalt; Crack500 dataset; The crack is located in the middle of the image; The crack is approximately 10 inches long; The crack is a thin, jagged line that runs horizontally across the image; The background of the image is grey asphalt;
\end{itemize}

\subsubsection{Prompts for \cref{fig:more_examples}}
\label{sec:more_examples_prompts}
\begin{itemize}
	\item \textbf{Crack500 (first row)}
	\begin{itemize}
		\item [(a)] This is an image of a crack in asphalt pavement; Crack500 dataset; The crack is located in the middle of the image; The crack is about 1.5 inches long and 0.5 inches wide; The crack is a straight line with a slight curve to the right 
		\item [(b)] This is a GT semantic map of a pavement crack image; Crack500 dataset; Semantic map centered in image; The crack is about 1.5 inches long and 0.5 inches wide; Pavement map reveals numerous vertical, horizontal, and diagonal cracks of varying widths and lengths
		\item [(c)] This is an image of a pavement crack; Crack500 dataset; The crack is located in the center of the image; The crack is about 10 cm long and 2 mm wide; The crack is a single, continuous crack that runs horizontally across the image
		\item [(d)] Crack500 dataset pavement crack GT map; Crack500 dataset; The semantic map is located in the center of the image; The crack is about 10 cm long and 2 mm wide; Map shows thick, tapered horizontal crack with slight upward curve
	\end{itemize}
	\item \textbf{Crack500 (second row)}
	\begin{itemize}
		\item [(a)] Asphalt pavement crack image; Crack500 dataset; The crack is located in the center of the image; 10-inch crack, 1/4-inch wide; The crack is linear and runs diagonally across the image 
		\item [(b)] GT semantic map of pavement crack image; Crack500 dataset; Semantic map in image center; 10-inch crack, 1/4-inch wide; Two main branches curve outward from central top in tree structure
		\item [(c)] This is an image of a pavement crack; Crack500 dataset; The crack is located in the middle of the image; The crack is approximately 10 inches long; Straight crack with minor deviations contains dark material
		\item [(d)] This is a semantic map (groundtrue (GT)) of a pavement crack image; Crack500 dataset; Map at image center; The crack is approximately 10 inches long; White 8-pixel wide curved semantic line from top left to bottom right
	\end{itemize}
	\item \textbf{XCAD}
	\begin{itemize}
		\item [(a)] Monochromatic arterial X-ray image with high contrast; XCAD dataset; Central, slightly right-skewed curvilinear structure; Vessels vary in size, with thicker mains tapering to finer branches; Arteries follow natural pathways with smooth to irregular shapes and complex branching
		\item [(b)] X-ray angiography GT map highlights curvilinear coronary arteries against solid background; XCAD dataset; Central map, clear periphery; Large semantic map with clear coronary arteries outline; Semantic map exhibits complex arterial structure with tapering, branching lines converging and diverging biomorphically
		\item [(c)] High-contrast monochrome X-ray of coronary arteries; XCAD dataset; Coronary angiography visible in center and upper left quadrant; Varies in thickness (2-4 mm) with multiple branches; Curvilinear structures exhibit a smooth, serpentine-like branching pattern with varying curve radii, resembling coronary vasculature
		\item [(d)] Semantic map of X-ray coronary angiography; XCAD dataset; Central map dominates image with sharp contrast on black background; Semantic map occupies most of the image, with variable width tapering at curvilinear structure ends; Vessels branch and taper, with smooth curves and divisions, including a sinuous main vessel with varying secondary vessels
	\end{itemize}
	\item \textbf{DCA1}
	\begin{itemize}
		\item [(a)] Monochromatic chest X-ray showing coronary arteries; DCA1 dataset; Coronary angiography is centrally located; Vessel diameter 1-4 mm, variable lengths; Arterial branching pattern exhibits smooth contours with no obstructions or narrowing, exhibiting diameter tapering
		\item [(b)] Coronary artery semantic map in black and white serves as ground truth for medical imaging; DCA1 dataset; Semantic map occupies upper two-thirds of image, centered vertically; The semantic map is of moderate size, not extending to the image edges; Complex vascular structure with multiple vessel bifurcations, originating top-left and branching towards bottom-right, resembling coronary artery paths
		\item [(c)] Chest radiograph of coronary arteries with contrast; DCA1 dataset; Coronary angiogram shows the heart's left coronary arteries; Vessel calibers range from fine to medium; Coronary arteries exhibit a curved, branching pattern; left anterior descending extends downwards, circumflex branches sideways
		\item [(d)] X-ray coronary angiography semantic map (GT); DCA1 dataset; Black-backed semantic map dominates image space; Semantic map spans image diagonally, from top left to bottom right; Complex map trend mimics coronary artery anatomy, with branching and curvilinear shapes forming acute angles at bifurcations
	\end{itemize}
	\item \textbf{CHUAC}
	\begin{itemize}
		\item [(a)] Coronary angiogram X-ray; CHUAC dataset; Central coronary angiogram predominantly in middle third; Angiography shows vessels of varying sizes; primary vessel larger, secondary branches smaller; Arteries branch curvilinearly, tapering from larger to smaller calibers toward the periphery
		\item [(b)] GT semantic map of coronary angiography image; CHUAC dataset; Black-backed semantic map central in image; Semantic map size varies, with line thicknesses indicating vessel calibers; Branched coronary arteries with smooth curves and sharp angles
		\item [(c)] A grayscale X-ray coronary angiography image; CHUAC dataset; Central coronary angiography highlights arteries against a low-density background; Coronary arteries vary in size, with the main branch several millimeters wide; Tortuous vasculature with smooth and irregular contours; curvilinear arteries with bifurcations and branching
		\item [(d)] BW semantic map of X-ray coronary angiography, delineating arteries from background; CHUAC dataset; Semantic map fills most of image space; Semantic map covers 60-70\% of image; Complex branching coronary arterial tree with varying thickness
	\end{itemize}
	\item \textbf{DRIVE}
	\begin{itemize}
		\item [(a)] Retinal fundus image with optic disc, macula, and vasculature; DRIVE dataset; Curvilinear vessels centered around optic disc, concentrated centrally and fanning peripherally; Central vessels taper into arterioles and venules ranging from micrometers to hundreds of micrometers; Vessels branch like trees, curving gently with varying tortuosity in smaller branches
		\item [(b)] BW semantic map of retinal blood network (GT image); DRIVE dataset; Centered semantic map occupies square image; denser veins near center, thinning outwards; Semantic map size matches image size, which is square; Branching vessels radiate from the optic disc, varying in thickness and forming networks
		\item [(c)] High-resolution retinal image shows vascular structures and surrounding tissue; DRIVE dataset; Blood vessels radiate outward from the optic disc, covering the retina; Vessel calibers range from <100  micrometer to 100-180  micrometer (arterioles, venules); Diverging tree-like vascular network with tortuous arteries and veins tapers from disc to periphery
		\item [(d)] GT retinal vessel map; DRIVE dataset; Central GT dominates canvas with ample margins; High-resolution fundus map covers entire field with varying vessel thicknesses; Semantic map reveals retinal vascular tree with central optic disc, vessels bifurcating to form dense network, thicker vessels near optic disc, finer branches toward periphery, arborized pattern typical of retinal vasculature
	\end{itemize}
	\item \textbf{CHASEDB1}
	\begin{itemize}
		\item [(a)] Fundus photo of eye interior (retina, vessels, optic disc, macula); CHASEDB1 dataset; Curvilinear vessels radiate from the optic disc to the peripheral retina; Finer vessels branch and curve away from the optic disc; Vasculature bifurcates centrally to peripherally with diminishing size; arteriole-venule crossings noted
		\item [(b)] Image of GT retinal vessel network; CHASEDB1 dataset; Central semantic map with radiating vessel structures; Semantic map covers entire image; Complex network with both thick and thin curvilinear structures, branching patterns, and bifurcation features
		\item [(c)] High-resolution image of the human retina highlighting its vascular network; CHASEDB1 dataset; Curvilinear retinal vessels emerge from the optic disc; Vessel caliber varies: central vessels prominent, peripheral vessels finer and sparser; Tortuous vessel paths, with larger vessels straighter and smaller vessels gently curving, form a fractal network
		\item [(d)] A black and white semantic map depicting the detailed structure of retinal vessels; CHASEDB1 dataset; Semantic map in image center with vessels extending outward; Semantic map size matches image resolution, showing detailed structure throughout; Tree-like semantic map branches from a center with varying vessel thickness and curvature
	\end{itemize}
\end{itemize}
\end{document}